\documentclass[runningheads]{llncs}

\usepackage[mobile]{eccv}

\usepackage{eccvabbrv}

\usepackage{graphicx}
\usepackage{booktabs}

\usepackage[linesnumbered, ruled]{algorithm2e}
\usepackage{bm}
\usepackage{times}
\usepackage{epsfig}
\usepackage{graphicx}
\usepackage{amsmath}
\usepackage{amssymb}
\usepackage{multirow}
\usepackage{booktabs}
\usepackage{wrapfig}  
\usepackage{color}
\usepackage{xcolor}
\usepackage{colortbl}
\SetKwRepeat{Do}{do}{while}%
\usepackage{makecell}
\usepackage{algorithmic}
\usepackage{multicol}
\usepackage{diagbox}

\newcommand{\mathvec}[1]{\mathbf{#1}}
\newcommand{\mathmat}[1]{\mathbf{#1}}

\newcommand{\mathimg}[1]{\mathbf{#1}}

\newcommand{\figref}[1]{Fig.~\ref{#1}}
\newcommand{\tabref}[1]{Table~\ref{#1}}
\newcommand{\secref}[1]{Section~\ref{#1}}
\newcommand{\algref}[1]{Algorithm~\ref{#1}}
\newcommand{\equref}[1]{Eq.~(\ref{#1})}

\newcommand{\tbdline}{\noindent\textcolor{red}{TBD TBD TBD TBD TBD TBD TBD TBD TBD TBD.}\\}
\newcommand{\tbd}[1][1]{ \newcount\tmp \tmp=0 \loop \advance\tmp by 1 \tbdline \ifnum\tmp<#1 \repeat }

\newcommand{\myparaheadd}[1]{\noindent\textbf{#1}}

\newcommand{\ourmethod}{DiffRetouch}
\newcommand{\best}[1]{\textbf{\color{red}{#1}}}
\newcommand{\secondbest}[1]{\textbf{\color{blue}{#1}}}

\newcommand{\markerlessfootnote}[1]{\begingroup
\renewcommand\thefootnote{}\footnote{#1}
\addtocounter{footnote}{-1}
\endgroup
}
\makeatletter
\newcommand\figcaption{\def\@captype{figure}\caption}
\newcommand\tabcaption{\def\@captype{table}\caption}
\makeatother
\usepackage[accsupp]{axessibility}  

\usepackage[pagebackref,breaklinks,colorlinks,citecolor=eccvblue]{hyperref}

\usepackage{orcidlink}

\begin{document}

\title{DiffRetouch: Using Diffusion to Retouch \\ on the Shoulder of Experts} 

\titlerunning{DiffRetouch: Using Diffusion to Retouch on the Shoulder of Experts}

\author{Zheng-Peng Duan\inst{1,2*} \and
Jiawei Zhang\inst{2} \and
Zheng Lin\inst{3} \and 
Xin Jin\inst{1} \and 
Dongqing Zou\inst{2,4} \and
Chunle Guo\inst{1} \and
Chongyi Li\inst{1\dag}
}

\authorrunning{Z.~Duan et al.}

\institute{VCIP, CS, Nankai University\and
SenseTime Research\and
BNRist, Department of Computer Science and Technology, Tsinghua University \and
PBVR
}

\maketitle
\vspace{-8mm}
\begin{abstract}
Image retouching aims to enhance the visual quality of photos.
Considering the different aesthetic preferences of users,
the target of retouching is subjective.
However, current retouching methods mostly adopt deterministic models,
which not only neglects the style diversity in the expert-retouched results and tends to learn an average style during training,
but also lacks sample diversity during inference.
In this paper,
we propose a diffusion-based method, named \ourmethod.
Thanks to the excellent distribution modeling ability of diffusion,
our method can capture the complex fine-retouched distribution covering various visual-pleasing styles in the training data.
Moreover, four image attributes are made adjustable to provide a user-friendly editing mechanism.
By adjusting these attributes in specified ranges,
users are allowed to customize preferred styles within the learned fine-retouched distribution.
Additionally, the affine bilateral grid and contrastive learning scheme are introduced to handle the problem of texture distortion and control insensitivity respectively.
Extensive experiments have demonstrated the superior performance of our method on visually appealing and sample diversity.
The code will be made available to the community.
\keywords{Image retouching \and Diffusion model}
\end{abstract}

\markerlessfootnote{* This work was done during an intership at Sensetime Research.}
\markerlessfootnote{\dag~Corresponding author.}

\vspace{-13mm}
\section{Introduction}
\label{sec:intro}
With the popularization of smartphones, 
taking photographs has become a daily activity for the public.
%
However,
captured photographs may be unsatisfactory due to varying factors like the illumination condition.
Thus, post-processing is inevitably desired.
A series of professional image-processing software
provide the users with useful tools to improve image quality.
However, 
these manual adjustments require specialized skills.
%
To help non-experts get visual-pleasing photos automatically,
numerous deep learning-based methods~\cite{moran2021curl,gharbi2017deep,wang2019underexposed} for image retouching have been proposed.

%
%
Considering the different aesthetic preferences of users,
retouching is a subjective process and the target style varies.
Even the same expert may adjust the images with different styles to satisfy various demands~\cite{song2021starenhancer}.
%
%
%
However,
most methods~\cite{he2020conditional,gharbi2017deep,zeng2020learning,moran2020deeplpf,moran2021curl} ignore the subjectivity of this task and adopt deterministic models.
Their drawbacks come from three aspects.
%
%
%
%
\textbf{1) Although trained with the subset retouched by one specific expert,
they neglect the intrinsic diversity within it, and actually learn the average style.}
This situation is more serious when trained with images retouched by multiple experts.
\textbf{2) During inference,
they can only produce one retouching style,
which may not always meet users' aesthetic preferences.}
%
%
To generate additional retouching results,
they need to train multiple models, 
which limits practical applications.
%
%
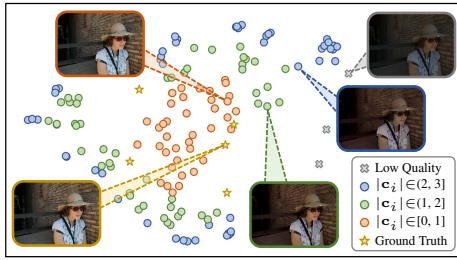
\begin{wrapfigure}{t!}{0.54\linewidth}
    \centering
    \includegraphics[width=0.93\linewidth]{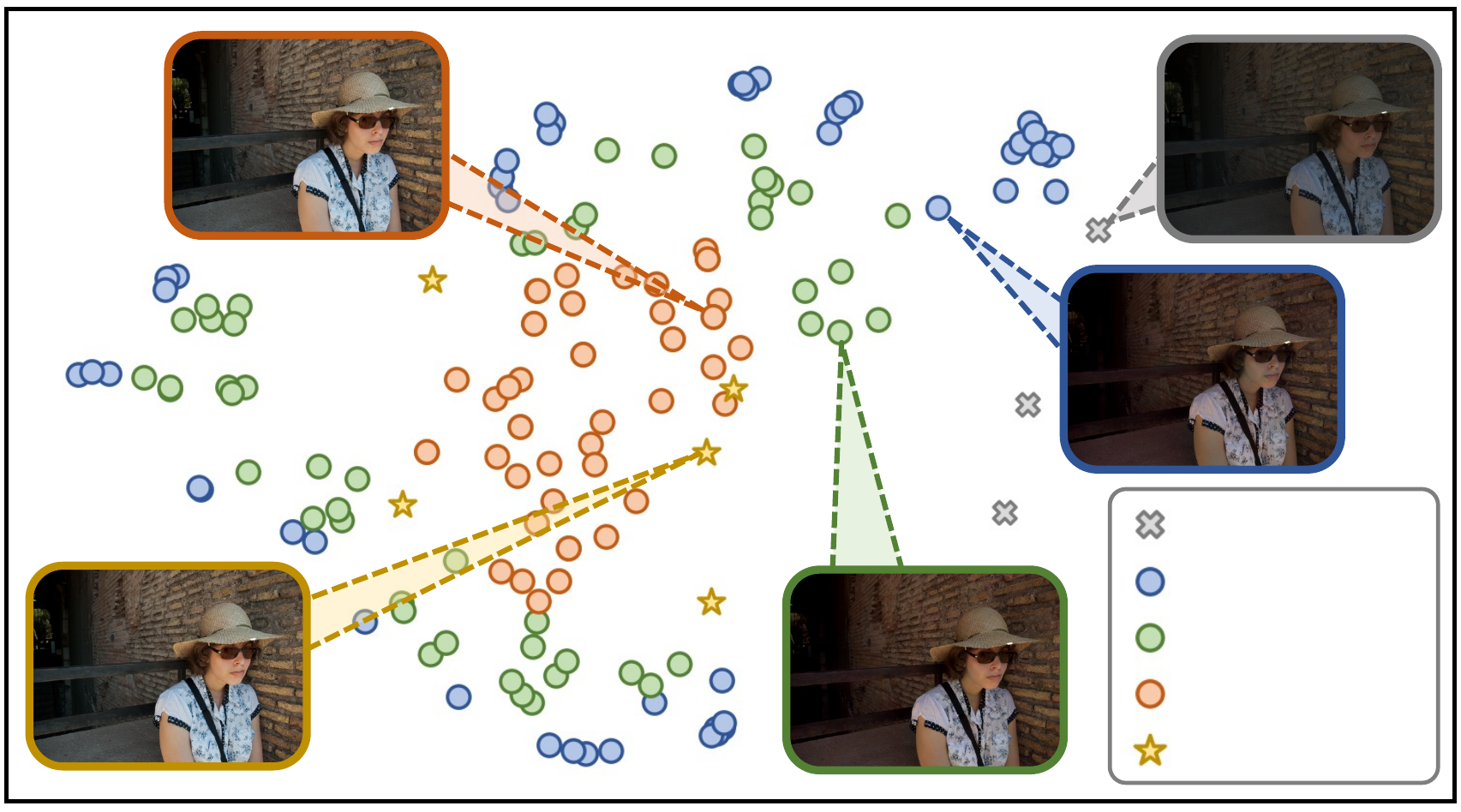}
    \put(-33.0, 33){\tiny{Low Quality}}
    \put(-33.0, 27){\tiny{$|\mathvec{c}_i|$$\in$(2, 3]}}
    \put(-33.0, 20){\tiny{$|\mathvec{c}_i|$$\in$(1, 2]}}
    \put(-33.0, 13){\tiny{$|\mathvec{c}_i|$$\in$[0, 1]}}
    \put(-33.0, 5.5){\tiny{Ground Truth}}
    \caption{
    \ourmethod~supports editing the retouching style by adjusting the condition $\mathvec{c}$, where each coefficient $\mathvec{c}_i$ corresponds to one image attribute.
    We generate numerous results with $|\mathvec{c}_i|$ randomly sampled in [0,1], (1,2], and (2,3].
    The features of these results extracted by style encoder~\cite{song2021starenhancer} are shown using t-SNE~\cite{van2008visualizing}.
    Since our \ourmethod~is trained with $|\mathvec{c}_i|$ limited to [0,1], the results sampled in this range are within the fine-retouched distribution surrounded by ground truths, otherwise, the results will deviate from it and be closer to low-quality images.
    This means that users can adjust within [0,1] to obtain preferred styles and meanwhile the final outputs tend to be objectively visual-pleasing.
    }
    \label{fig:teaser}
    \vspace{-8mm}
\end{wrapfigure}
\textbf{3) Although several methods~\cite{song2021starenhancer,kim2020pienet} support additional styles,
extra images are required to indicate the desired style.}
This adds extra burden to the users and adjusting the style through images is ambiguous.
%
%

In practice, a retouching method needs to cover the fine-retouched distribution,
which includes various visual-pleasing styles to satisfy different aesthetic preferences.
%
Recently,
diffusion~\cite{ho2020denoising, nichol2021improved} has shown its strong modeling ability for complex distributions.
%
%
%
In this work,
we introduce the diffusion model into image retouching,
%
where the benefits come from two aspects.
1) 
%
%
The diffusion-based model can capture the complex distributions covering various styles appearing in the training data even when trained with results retouched by multiple experts.
%
%
%
%
2) 
During inference,
the model can generate various styles within the fine-retouched distribution 
with no additional images.

More specifically,
we propose a Stable Diffusion-based~\cite{rombach2022high} retouching method,
%
which is trained by directly conditioning on low-quality input images via concatenation.
In order to provide the user with a friendly and understandable editing mechanism,
four image attributes (colorfulness, contrast, color temperature, and brightness) are made adjustable by coefficients.
These coefficients constitute a vector and are then mapped to the intermediate layers of the U-Net via a cross-attention mechanism.
However,
due to the information loss existing in the encoding and decoding process~\cite{jiang2023autodir},
the results will exhibit noticeable distortion in image textures,
which we refer to as \textbf{texture distortion}.
%
Inspired by HDRNet~\cite{gharbi2017deep},
we introduce the affine bilateral grid to address the problem.
Besides the latent prediction for the progressive denoising process,
the underlying U-Net backbone within each denoising step outputs the affine bilateral grid as well.
As for the last denoising step,
we directly apply the affine transformations, which are inferred from the obtained bilateral grid, to the input image.
%
%
The other problem is that the influence caused by adjusting these attributes tends to be weak to satisfy practical needs,
which we denote as \textbf{control insensitivity}.
To encourage models more aware of the adjustment brought by each coefficient,
we design the contrastive learning scheme,
which involves explicit supervision \wrt these attributes.
%
It is worth mentioning that regardless of how the coefficients are adjusted within the specified range, 
the final result tends to be sampled from the learned fine-tuned distribution,
which is visualized in \figref{fig:teaser}.
%
That is to say, users are allowed to adjust within the specified range to obtain preferred styles, 
and meanwhile the final outputs are likely to be objectively visual-pleasing results.

Our contributions can be summarized as follows:
\begin{itemize}
    \item We propose a diffusion-based retouching method, to cover the fine-retouched distribution, along with four adjustable attributes to edit the final results.
    \item The bilateral grid is introduced into the diffusion model to overcome the texture distortion caused by the information loss in the encoding and decoding process.
    \item To address the control insensitivity, we design the contrastive learning scheme to encourage models more aware of the adjustment brought by each coefficient.
\end{itemize}

\vspace{-5mm}
\section{Related Work}
\label{sec:related work}
\myparaheadd{Deep Image Retouching.}
With the datasets~\cite{bychkovsky2011learning, liang2021ppr10k} proposed for image retouching,
deep learning-based methods have made impressive progress.
Many methods~\cite{chen2018learning,kim2021representative,sun2021enhance,chen2018deep,he2020conditional} utilize fully convolutional networks to perform an image-to-image translation.
Another research line is to combine deep learning with physical models.
These methods use neural networks to fit physical models such as 
Retinex theory~\cite{wang2019underexposed, zhu2020zero, liu2021retinex},
3D-LUT~\cite{zeng2020learning, yang2022adaint, wang2021real}, 
curve-based color transformations~\cite{moran2021curl,li2020flexible,li2021learning,li2022cudi,song2021starenhancer},
affine bilateral grid~\cite{gharbi2017deep},
and parametric filters~\cite{moran2020deeplpf}.
Among them,
only a few methods~\cite{song2021starenhancer,kim2020pienet,kim2023learning} support additional styles.
%
However, extra images are required to indicate the desired style,
which is inconvenient to users.
%
%
Some methods~\cite{hu2018exposure,kosugi2020unpaired,ouyang2023rsfnet,Tseng2022NeuralPhotoFinishing} decouple the task into a series of operations utilized in professional software,
thus offering understandable options to edit the results.
Still limited by their deterministic networks,
these methods can only generate one automatic retouching result,
and subsequent adjustments challenge the aesthetics of users to guarantee the final results are aesthetically pleasing.
%
%
Our \ourmethod~maintains their advantages and steps further.
We provide users with an understandable editing mechanism \wrt image attributes, and once adjusting in the specified range, users are allowed to customize preferred styles within the fine-retouched distribution.
Inspired by~\cite{wang2023exploring}, 
we select colorfulness, contrast, color temperature, and brightness as the four image attributes,
which can be further extended to other attributes.

\myparaheadd{Diffusion for Low-Level Vision.}
With the improvement of denoising diffusion probabilistic models (DDPM)~\cite{ho2020denoising,nichol2021improved},
diffusion model has shown potential in low-level vision tasks.
%
Some researchers adopt the framework of DDPM to handle specific tasks,
such as super-resolution~\cite{saharia2022image}, inpainting~\cite{lugmayr2022repaint, xie2023smartbrush},
deraining~\cite{ozdenizci2023restoring},
deblurring ~\cite{whang2022deblurring,ren2023multiscale},
low-light enhancement~\cite{wang2023exposurediffusion,yin2023cle,hou2023global},
colorization~\cite{saharia2022palette},
and shadow removal~\cite{guo2023shadowdiffusion}.
The other category of works~\cite{wang2022zero,kawar2022denoising,meng2022diffusion,chung2022improving,fei2023generative,luo2023image} use a pre-trained DDPM to realize unified image restoration without task-oriented training.
Recently,
with the popularity of Stable Diffusion~\cite{rombach2022high},
several methods~\cite{wang2023exploiting,lin2023diffbir,jiang2023autodir} are proposed to leverage its generative prior to cope with blind image restoration.
There are also some methods~\cite{yang2023pixel, wu2023seesr, chen2023image, sun2023coser, yu2024scaling} deeply explore the potential of Stable Diffusion in image super-resolution.
Our \ourmethod~belongs to the first category 
%
and aims to employ the excellent distribution coverage ability of diffusion to capture the complex fine-retouched distribution.

\begin{figure}[t]
    \centering
    \includegraphics[width=\linewidth]{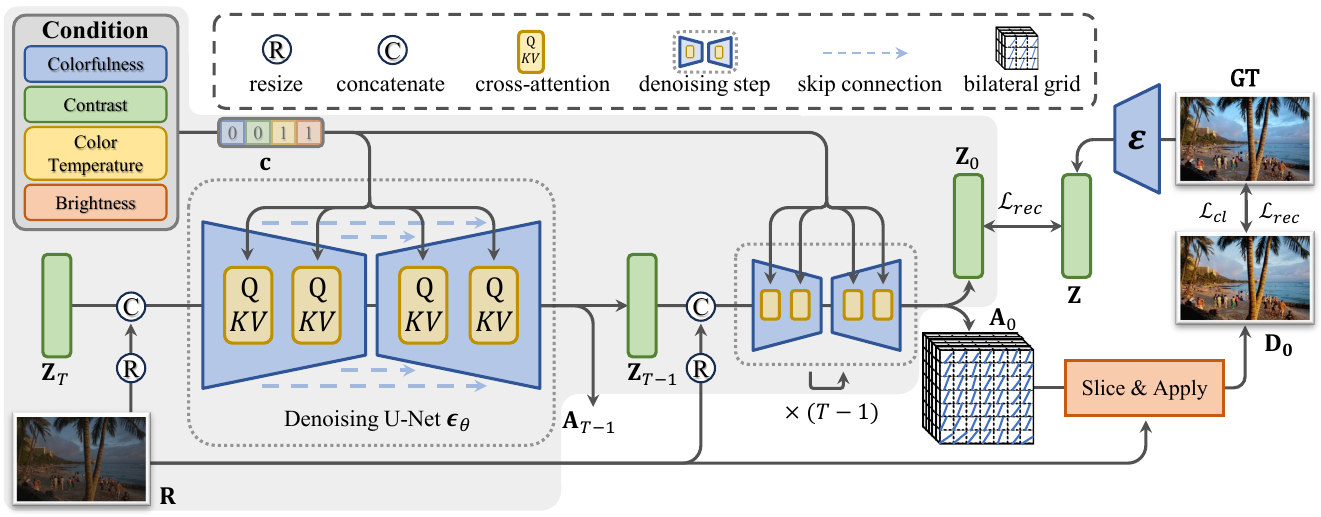}
    \vspace{-6mm}
    \caption{Pipeline of our \ourmethod. The sampling process and supervision during training are included. The baseline model part is marked in \textcolor{gray}{gray}. 
    The affine bilateral grid and $\mathcal{L}_{cl}$ are additionally introduced in DiffRetouch to tackle texture distortion and control insensitivity.
    During training, the denoising model takes the noisy latent $\mathimg{Z}_t$, resized version of $\mathimg{R}$ and condition $\mathvec{c}$ \wrt image attributes as input for each step, 
    then generates $\mathimg{Z}_{t-1}$ and affine bilateral grid $\mathimg{A}_{t-1}$ simultaneously.
    After looking up in $\mathimg{A}$ based on the position and intensity of each pixel in $\mathimg{R}$, which is similar to~\cite{gharbi2017deep}, the output $\mathimg{D}$ is obtained by matrix multiply between the sliced affine matrics and pixel colors of $\mathimg{R}$.
    $\mathcal{L}_{rec}$ (\equref{eq:Lrec}) is imposed in both the latent ($\mathimg{Z}$) and pixel ($\mathimg{D}$) space, along with the $\mathcal{L}_{cl}$ (\equref{eq:Lcl}).
    During inference, at each step of the sampling, $\mathimg{Z}_{t-1}$ is used as the input of the next denoising step for the progressive denoising process.
    Only for the last step, $\mathimg{A}_{0}$ is used to obtain the final output $\mathimg{D}_0$.
    }
    \label{fig:pipeline}
    \vspace{-6mm}
\end{figure}

\section{Methodology}
\label{sec:method}
\subsection{Overview}
Given an image $\mathimg{R}$ suffering from photographic defects,
such as improper exposures and limited contrast,
image retouching aims to generate a visually pleasing rendition.
%
%
%
%
%
%
In this work, we propose the Stable Diffusion-based method for retouching, named \ourmethod.
%
%
%
%
To help users customize the styles that better fit their aesthetic preferences,
%
four image attributes (colorfulness, contrast, color temperature, and brightness)
are made adjustable by coefficients and constitute the condition input $\mathvec{c}$.
Note that our method is a framework where these attributes are extendable.
Once the score of another attribute can be calculated, that attribute control can be introduced.
%
%
The direct application of Stable Diffusion, along with the condition inputs ($\mathimg{R}$ and $\mathvec{c}$), is viewed as the baseline model,
which is introduced in \secref{sec:baseline} and marked in \textcolor{gray}{gray} in \figref{fig:pipeline}.
However, the information loss existing in the encoder and decoder process causes texture distortion.
To overcome the texture distortion,
%
we introduce the affine bilateral grid into our baseline model, 
%
which is detailed in \secref{sec:hdrnet}.
%
The implementation of the affine bilateral grid is in \textcolor{violet}{violet} in \algref{alg:training} and \algref{alg:sampling}.
%
%

Stable Diffusion~\cite{rombach2022high} adopts the DDPM training strategy in the latent space,
%
%
where the reconstruction supervision is imposed on the latent for predicting the added noise.
%
For the training of affine bilateral grid,
the reconstruction supervision is also utilized in the pixel space.
These two supervisions are defined in \secref{sec:reconstruct}.
%
To alleviate the control insensitivity,
we design the contrastive learning scheme, which is introduced in \secref{sec:contrastive}.
The part \wrt contrastive learning is in \textcolor{orange}{orange} in \algref{alg:training}.
%
%
%
%
%

%


%
\vspace{-4mm}
\subsection{Architecture}
\label{sec:architecture}

\subsubsection{Baseline}
\label{sec:baseline}

The baseline model of our \ourmethod~is built upon Stable Diffusion~\cite{rombach2022high},
which is a variant of DDPM~\cite{ho2020denoising,chung2022improving}.
%
DDPM involves sequentially corrupting training data with noise, and then learning to reverse this corruption.
%
%
Specifically,
denoising model $\bm{\epsilon}_\theta(\mathimg{X}_t,t)$ is trained to predict the added noise on the sampled image $\mathimg{X}$,
where $\mathimg{X}_t$ is a noisy version of $\mathimg{X}$ at timestamp $t$.
The underlying backbone of the denoising model $\bm{\epsilon}_\theta$ is the time-conditional U-Net.
To improve both the training and sampling efficiency of DDPM,
Stable Diffusion applies the forward and reverse processes in the latent space rather than in pixel space.
Equipped with powerful pre-trained autoencoders consisting of encoder $\mathcal{E}$ and decoder $\mathcal{D}$,
Stable Diffusion can efficiently obtain the latent space representation $\mathimg{Z}$ of $\mathimg{X}$ by $\mathimg{Z} = \mathcal{E}(\mathimg{X})$ during training,
and transform the latent space samples to the pixel space through $\mathcal{D}$.
%
%
Moreover,
by replacing the original denoising model with the conditional one $\bm{\epsilon}_\theta(\mathimg{Z}_t,t, \mathvec{m})$, where  $\mathimg{Z}_t$ is the noisy latent at timestamp $t$ and $\mathvec{m}$ is the condition input such as text,
Stable Diffusion can be turned into more flexible conditional image generators.
Our baseline also performs the DDPM process in the latent space and utilizes the conditional denoising models.
The sampling process starts from the Gaussian noise map $\mathimg{Z}_T$,
where $T$ is the total number of time steps for denoising,
and $\mathimg{Z}_T$ has a smaller resolution than input image $\mathimg{R}$.
%
%
In order to provide the information of the image to be retouched,
$\mathimg{R}$ is resized to the same resolution as $\mathimg{Z}_t$,
and the resized image $\mathimg{R}'$ is fed into the denoising model via concatenation with $\mathimg{Z}_t$.
Another input condition is a vector consisting of four adjustable coefficients regarding four attributes (colorfulness, contrast, color temperature, and brightness),
which can be denoted as $\mathvec{c} = [\mathvec{c}_1,\mathvec{c}_2,\mathvec{c}_3,\mathvec{c}_4]$.
%
%
In this design,
by adjusting $\mathvec{c}_i$,
the final retouching style of the output can be edited on the corresponding attributes.
%
Once the coefficient $\mathvec{c}_i$ is adjusted within $[-1,1]$,
the final result is likely to be sampled in the learned fine-retouched distribution.
%
%
For example, 
when $\mathvec{c}_i$ is set to 1,
this will make the output more colorful, higher contrast, cold temperature, or brighter one among the fine-retouched results,
and vice versa.
%
%
%
The adjustable vector is mapped to the intermediate layers of the backbone U-Net via a cross-attention layer,
where the details can be found in the supplementary material.
%
%

For each denoising step during sampling, 
the denoising model takes the noisy latent $\mathimg{Z}_t$, the resized low-quality image $\mathimg{R}'$, and adjustable vector  $\mathvec{c}$ as input, and output the noise prediction, which can be formulated as
\begin{equation}
\label{eq:tradition}
    \bm{\epsilon}_{pred,t} = \bm{\epsilon}_\theta\left([\mathimg{Z}_{t}, \mathimg{R}'], t, \mathvec{c}\right).
\end{equation}
Following \algref{alg:sampling}-6,
we can obtain $\mathimg{Z}_{t-1}$ with the noise prediction,
which is the input for the next denoising step.
The denoising step repeats until obtaining $\mathimg{Z}_0$.
By passing $\mathimg{Z}_0$ through $\mathcal{D}$,
we can obtain the result output by the baseline model.

\begin{figure}[tb]
  \centering
  \begin{minipage}{0.48\linewidth}
    \includegraphics[width=\linewidth]{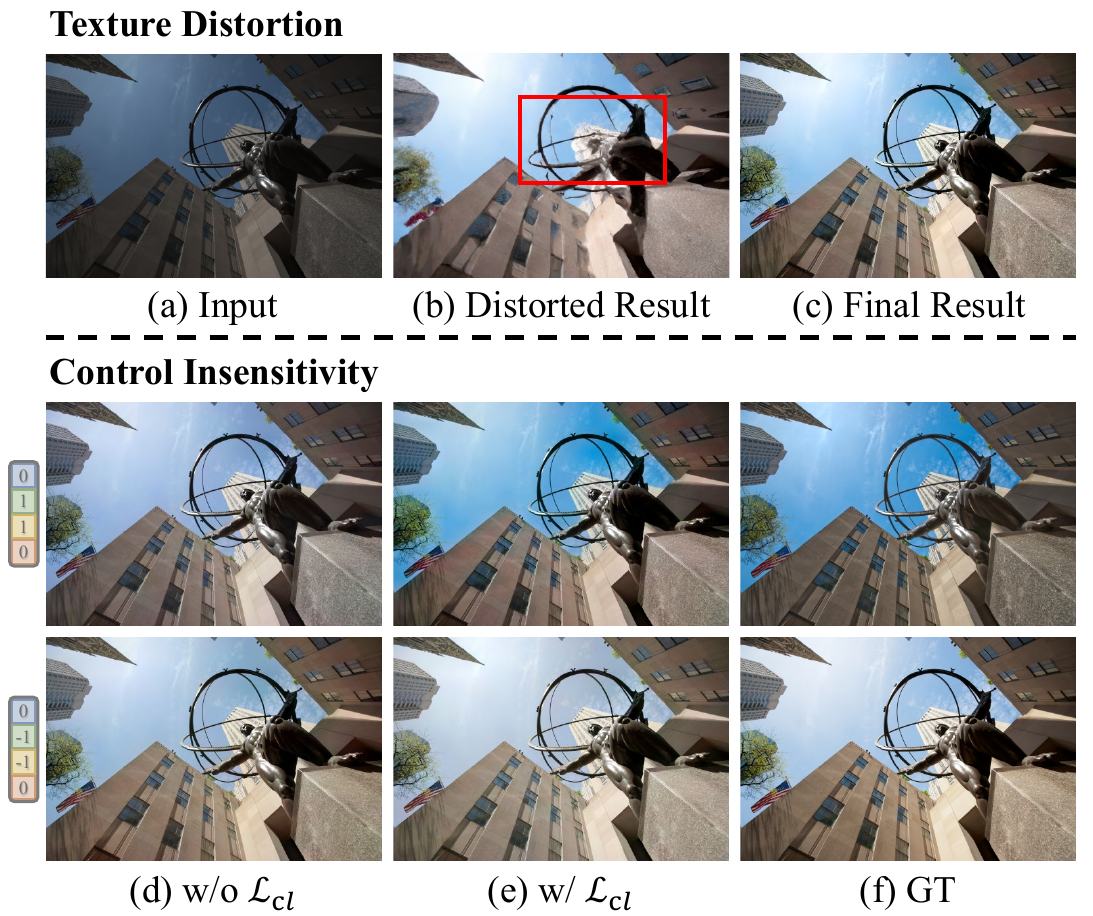}
    
    \caption{
    Examples of \textbf{Texture Distortion} and \textbf{Control Insensitivity}.
    The top row: (a) Input image; (b) and (c) are the results generated w/o and w/  the affine bilateral grid.
    The bottom two rows: (d) and (e) are the results generated by the model w/o and w/ $\mathcal{L}_{cl}$;
    (f) are the results retouched by two experts as GT.
    The input condition $\mathvec{c}$ is shown on the left,
    where the adjusted attributes are contrast and color temperature.
    With $\mathcal{L}_{cl}$ (\equref{eq:Lcl}), the region of the sky is closer to the expert-retouched results.}
    \label{fig:problem}
  \end{minipage}
  \hfill
  \begin{minipage}{0.48\linewidth}
    \includegraphics[width=\linewidth]{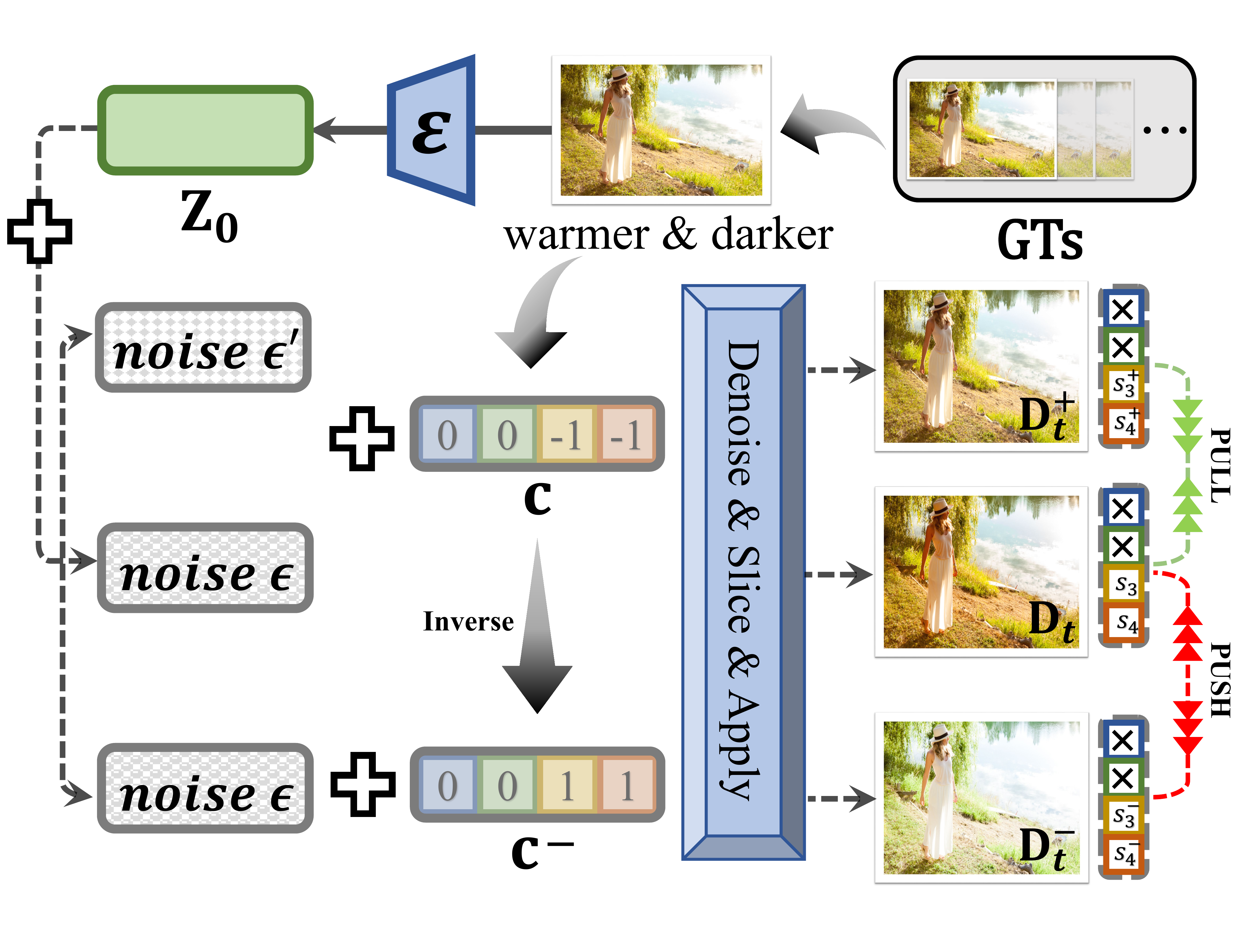}
    \caption{Framework of contrastive learning scheme.
    The regular branch takes the latent $\mathimg{Z}_0$, the noise map $\bm{\epsilon}$, and the condition $\mathvec{c}$ as input to generate the result $\mathimg{D}_t$.
    Another two branches produce the positive sample $\mathimg{D}^+_t$ with a different noise map $\bm{\epsilon'}$ and the same condition $\mathvec{c}$,
    and negative samples $\mathimg{D}^-_t$ with the same $\bm{\epsilon}$ and the opposite condition $\mathvec{c}^-$.
    For coefficients $\mathvec{c}_i \neq 0$, $\mathcal{L}_{cl}$  (\equref{eq:Lcl}) steers the corresponding $\mathvec{s}_i$ closer to $\mathvec{s}^+_i$ and away from $\mathvec{s}^-_i$.
    In this example, the adjusted attributes are color temperature and brightness.
   }
    \label{fig:contrast}
  \end{minipage}
  \vspace{-6mm}
\end{figure}

\subsubsection{Affine Bilateral Grid}
\label{sec:hdrnet}
%


Despite the great power and good training of the pre-trained autoencoders,
the result output by the baseline model has severe texture distortion,
which is shown in the top row of \figref{fig:problem}.
%
%
The distortion is mainly caused by the information loss existing in both the encoding and decoding processes.
%
%

Inspired by HDRNet~\cite{gharbi2017deep},
we introduce the affine bilateral grid $\mathimg{A}$ into our \ourmethod~to overcome the texture distortion.
$\mathimg{A}$ can be viewed as a 3D array, where each grid is indexed by position and intensity.
%
In each grid,
$\mathimg{A}$ stores a $3 \times 4$ affine matrix.
When applying $\mathimg{A}$ on the full-resolution image,
we can lookup in $\mathimg{A}$ with the position and intensity of each pixel, 
and retrieve the affine matrix through trilinear interpolation.
Then the final color for each pixel is obtained by matrix multiply between the affine matrix and the original color.
%
%
Due to the nature of the bilateral grid,
nearby pixels with similar intensities tend to retrieve similar affine matrics.
%
After matrix multiplication, these nearby pixels still keep similar colors in the output, avoiding distortion.
%
%

\begin{algorithm}[ht]
    \caption{Training Process}
    \label{alg:training}
    \begin{multicols}{2}
    \begin{algorithmic}[1]
    \REPEAT
    \STATE $ \mathimg{X}_0, \mathimg{R}, \mathvec{c} \sim q\left(\mathimg{X}_0, \mathimg{R}, \mathvec{c}\right)$ \;
    \tcp{$q\left(\mathimg{X}_0, \mathimg{R}, \mathvec{c}\right)$ is the distribution of real data}
    \STATE $\mathimg{Z}_0 = \mathcal{E}\left(\mathimg{X}_0\right) $\;
    \STATE $t \sim \text {Uniform}\left(\{1,\dots,T\}\right)$ \;
    \STATE $\bm{\epsilon}\textcolor{orange}{,  \bm{\epsilon}^{'} }  \sim \mathcal{N}(\mathimg{0},\mathimg{I})$  \;
    \STATE $\mathimg{R}' = \text{Resize}\left(\mathimg{R}\right) $\;
    \tcp{$1-\bar{\alpha}_t$ is the noise schedule}
    \STATE $\mathimg{Z}_{t} = \sqrt{\bar{\alpha}_t} \mathimg{Z}_0 + \sqrt{1-\bar{\alpha}_t}\bm{\epsilon} $\;
    \STATE $\textcolor{orange}{\mathimg{Z}_{t}' = \sqrt{\bar{\alpha}_t} \mathimg{Z}_0 + \sqrt{1-\bar{\alpha}_t}\bm{\epsilon}^{'} } $\; 
    \STATE $[\bm{\epsilon}_{pred,t}\textcolor{violet}{, \tilde{\mathimg{A}}_t}] = \bm{\epsilon}_\theta\left([\mathimg{Z}_{t}, \mathimg{R}'], t, \mathvec{c}\right)$ \;
    \tcp{Positive Sample} 
    \STATE $\textcolor{orange}{[\bm{\epsilon}_{pred,t}^+, \tilde{\mathimg{A}}^+_t] = \bm{\epsilon}_\theta\left([\mathimg{Z}_{t}', \mathimg{R}'], t, \mathvec{c}\right)}$ \; 
    \tcp{Negative Sample} 
    \STATE $\textcolor{orange}{[\bm{\epsilon}_{pred,t}^-, \tilde{\mathimg{A}}^-_t] = \bm{\epsilon}_\theta\left([\mathimg{Z}_{t}, \mathimg{R}'], t, \mathvec{-c}\right)}$ \;
    \STATE $\textcolor{violet}{\mathimg{A}_t}\textcolor{orange}{/\mathimg{A}^+_t/ \mathimg{A}^-_t} \newline
    \textcolor{violet}{= \text{Reshape} (\tilde{\mathimg{A}}_t\textcolor{orange}{/\tilde{\mathimg{A}}_t^+/\tilde{\mathimg{A}}_t^-})}$ \;
    \STATE $\textcolor{violet}{\mathimg{D}_t}\textcolor{orange}{/\mathimg{D}^+_t/\mathimg{D}^-_t} \newline \textcolor{violet}{= \text{Slice\&Apply}(\mathimg{A}_t\textcolor{orange}{/\mathimg{A}^+_t/\mathimg{A}^-_t},\!\mathimg{R})}$\;
    \tcp{Calculate the scores \wrt four atrributes} 
    \STATE $\textcolor{orange}{\mathvec{s}/ \mathvec{s}^+/ \mathvec{s}^-} \newline \textcolor{orange}{ = \text{Calculate}(\mathimg{D}_t/ \mathimg{D}^+_t/\mathimg{D}^-_t)}$ \;
    \STATE Take gradient descent step on \newline
    $\quad \nabla_\theta( \Vert \bm{\epsilon}_{pred,t} - \bm{\epsilon} \Vert ^2 \textcolor{violet}{+ \beta \Vert \mathimg{D}_{t} - \mathimg{X}_0 \Vert ^2 } $ \\ \textcolor{orange}{+ $\mathcal{L}_{cl}$ })  \;
    \tcp{$\mathcal{L}_{cl}$ is formulated as \equref{eq:Lcl} .}
    \UNTIL{converged} \;
    \tcp{The part \wrt affine bilateral grid is in \textcolor{violet}{violet}. \\The part \wrt contrastive learning is in \textcolor{orange}{orange}.}
    \end{algorithmic}
    \end{multicols}
    \vspace{-2mm}
\end{algorithm}

In our implementation,
besides the noise estimation, the underlying U-Net outputs the affine bilateral grid as well,
which can be formulated as:
\begin{equation}
\label{eq:diffretouch}
    [\bm{\epsilon}_{pred}, \tilde{\mathimg{A}}] =\bm{\epsilon}_\theta\left([\mathimg{Z}_{t}, \mathimg{R}'], t, \mathvec{c}\right).
\end{equation}
%
%
After unrolling the channels of the output $\tilde{\mathimg{A}}$,
we can obtain the affine bilateral grid $\mathimg{A}  \in \mathbb{R}^ {H_{grid} \times W_{grid} \times D \times 12}$,
where $D$ is the dimension of intensity and 12 represents the $ 3 \times 4$ affine matrix.
Note that $H_{grid}$ and $W_{grid}$ are the height and width of $\mathimg{A}$,
which are smaller than the resolution of $\mathimg{R}\in \mathbb{R}^{ H\times W\times 3}$.
The operation of slicing and applying $\mathimg{A}$ on the full-resolution image is similar to HDRNet~\cite{gharbi2017deep}.
%
The guidance map $\mathimg{G} \in \mathbb{R}^{ H\times W\times 1}$ is obtained by feeding $\mathimg{R}$ into a pixel-level network.
%
%
Given the position and intensity indicated by $\mathimg{G}$, 
the slicing performs a per-pixel lookup in $\mathimg{A}$ and retrieve the sliced $\mathimg{A}' \in \mathbb{R}^{ H\times W\times 1\times 12}$ through trilinear interpolation.
The final result is obtained by applying the matrix multiply between the affine matrics and the original color for each pixel in $\mathimg{R}$.
%
%
%
%
%
As shown in the top row of  \figref{fig:problem},
the final result of our \ourmethod~maintains the original details.
%
%


    

\vspace{-4mm}
\subsection{Training Strategy}
\subsubsection{Reconstruction Supervision}
\label{sec:reconstruct}
Stable Diffusion performs the standard DDPM training strategy in the latent space.
More specifically,
with the time step $t$ randomly sampled from a uniform distribution,
the noisy latent $\mathimg{Z}_t$ at the timestamp $t$ can be obtained by \algref{alg:training}-7.
Then along with resized image $\mathimg{R}'$ and the condition $\mathvec{c}$ as input,
the denoising model is trained to predict the noise $\bm{\epsilon}$ according to \equref{eq:tradition}.
Since the clean latent $\mathimg{Z}_0$ can be estimated with the predicted noise $\bm{\epsilon}_{pred,t}$,
the supervision between $\bm{\epsilon}$ and $\bm{\epsilon}_{pred,t}$ can be viewed as the reconstruction supervision in the latent space.

\label{sec:training}

%
In our \ourmethod,
the role of the denoising model is to generate the noise estimation and the affine bilateral grid simultaneously, which is indicated by \equref{eq:diffretouch}.
For the training of the affine bilateral grid,
we also impose the reconstruction supervision on the pixel-space output $\mathimg{D}_t$.
We collectively refer to these two supervisions in both the latent ($\mathimg{Z}$) and pixel ($\mathimg{D}$) space as reconstruction supervision.

\vspace{-2mm}
\subsubsection{Contrastive Learning}
\label{sec:contrastive}
Before introducing our contrastive learning, a brief description of the construction of image-condition pairs is needed.
Current datasets for retouching consist of results retouched by different experts.
However, 
they lack a description of the retouching style for each result.
%
%
%
In order to provide the users with an understandable editing mechanism,
we describe the retouching style from the aspects of four image attributes  (colorfulness, contrast, color temperature, and brightness).
%
To construct such image-condition pairs,
commonly-used measurements are utilized to calculate the score for each attribute,
%
which can be denoted as $\mathvec{s} = [\mathvec{s}_1,\mathvec{s}_2,\mathvec{s}_3,\mathvec{s}_4]$.
%
For each low-quality input $\mathimg{R}$,
the dataset provides several ground truth (GT) retouched by different experts.
%
%
Among these GTs for the same $\mathimg{R}$,
we first calculate their $\mathvec{s}$ and initialize their $\mathvec{c}$ as 0.
%
Then for each attribute ($i$),
the corresponding coefficient $\mathvec{c}_i$ is set to 1(-1) for the GT with the highest(lowest) $\mathvec{s}_i$. 
That is to say, for the most colorful, highest contrast, coolest, and brightest GT among the GTs for the same $\mathimg{R}$, the corresponding coefficient $\mathvec{c}_i$ will be set as 1, and vice versa.
The details about the score calculation and condition construction can be found in the supplementary material.
Note that when used in practice, these conditions are provided and adjusted by the users according to their preferences.

\begin{algorithm}[t!]
    \renewcommand{\algorithmicrequire}{ \textbf{Input:}} 
    \renewcommand{\algorithmicensure}{ \textbf{Output:}} 
    \caption{Sampling Process}
    \label{alg:sampling}
    \begin{multicols}{2}
    \begin{algorithmic}[1]
    \REQUIRE{Low-quality Image $\mathimg{R}$, Condition $\mathvec{c}$} \;
    \ENSURE{Retouched Result $\mathimg{D}_0$}\;
    \STATE $\mathimg{Z}_T \sim \mathcal{N}(\mathimg{0},\mathimg{I})$ \;
    \STATE $\mathimg{R}' = \text{Resize}(\mathimg{R})$\;
    \FOR{$t=T, T-1,\dots,1$}
    \STATE$\mathvec{z} \sim \mathcal{N}(\mathimg{0},\mathimg{I})$ if $t>0$, else $\mathvec{z}=\mathimg{0}$ \;
    \STATE$[\bm{\epsilon}_{pred,t} \textcolor{violet}{, \tilde{\mathimg{A}}_{t-1}}] = \bm{\epsilon}_\theta([\mathimg{Z}_{t}, \mathimg{R}'], t, \mathvec{c})$ \;
    \STATE$\mathimg{Z}_{t-1} = \frac{1}{\sqrt{\alpha_t}}(\mathimg{Z}_t - \frac{1-\alpha_t}{\sqrt{1-\bar{\alpha}_t}}\bm{\epsilon}_{pred,t}) + \sigma_t\mathvec{z}$ \;
    \ENDFOR
    \STATE$\textcolor{violet}{\mathimg{A}_0 = \text{Reshape} (\tilde{\mathimg{A}}_{0})}$\;
    \STATE$\textcolor{violet}{\mathimg{D}_0 = \text{Slice} \& \text{Apply} (\mathimg{A}_0, \mathimg{R})}$ \;
    \STATE \Return $\mathimg{D}_0$ \;
    \tcp{The part \wrt affine bilateral grid is in \textcolor{violet}{violet}}
    \end{algorithmic}
    \end{multicols}
\vspace{-2mm}
\end{algorithm}
Although training with such image-condition pairs under the reconstruction supervision enables the model to sample various styles according to $\mathvec{c}$, 
the influence caused by adjusting these coefficients tends to be weak.
As shown in \figref{fig:problem} (d),
adjusting the coefficients related to contrast and color temperature has little effect on the final result.
In order to encourage our \ourmethod~more aware of the adjustment brought by each attribute,
we design the contrastive learning scheme,
which involves explicit supervision \wrt these attributes.
%
An example is shown in \figref{fig:contrast}.

More specifically,
%
apart from the regular branch which takes the latent $\mathimg{Z}_0$, the noise map $\bm{\epsilon}$, and the condition $\mathvec{c}$ as input and operates as \equref{eq:diffretouch},
another two branches are included to produce the positive and negative samples.
The positive branch shares the same $\mathvec{c}$, 
but adopt another noisy map $\bm{\epsilon}'$ as input.
The negative branch still utilizes $\bm{\epsilon}$ but with the opposite condition $\mathvec{c}^-=-\mathvec{c}$.
%
After the operation of slicing and applying,
three results in the pixel space can be obtained,
then we calculate their scores \wrt the four attributes,
which can be denoted as $\mathvec{s}, \mathvec{s}^+,$ and $ \mathvec{s}^-$ respectively.
For the attribute whose coefficient is not 0,
we steer the corresponding score $\mathvec{s}_i$ closer to $\mathvec{s}_i^+$ and away from $\mathvec{s}_i^-$.
The effect of the contrastive learning scheme is shown in \figref{fig:problem}.
Equipped with contrastive learning, the influence brought by the condition input is more sensitive and the results are closer to that retouched by experts.

So far, the overall training objective can be written as
\begin{equation}
    \mathcal{L} = \mathcal{L}_{rec} + \lambda \mathcal{L}_{cl}.
\end{equation}
The reconstruction supervisions are imposed on both the latent and pixel space,
which can be formulated as
\begin{equation}
    \label{eq:Lrec}
    \mathcal{L}_{rec} = \Vert \bm{\epsilon}_{pred,t} - \bm{\epsilon} \Vert ^2 + \beta \Vert \mathimg{D}_{t} - \mathimg{X}_0 \Vert ^2,
\end{equation}
where $\lambda$ and $\beta$ are the scalars.
Inspired by InfoNCE~\cite{oord2018representation},
the loss function of the contrastive learning is defined as
\begin{equation}
    \label{eq:Lcl}
    \mathcal{L}_{cl}\!=\!\sum_{\{i|\mathvec{c}_i \neq 0\}}\!\!-\log \frac{e^{-|\mathvec{s}_i - \mathvec{s}_i^+|/\tau}}{e^{-|\mathvec{s}_i - \mathvec{s}_i^+|/\tau}\!+\!e^{-|\mathvec{s}_i - \mathvec{s}_i^-|/\tau}},
\end{equation}
where $\tau$ is the temperature parameter.
$\mathvec{s}_i$, $\mathvec{s}_i^+$, and $\mathvec{s}_i^-$ are the scores of the pixel space output $\mathimg{D}_t$, $\mathimg{D}_t^+$, and $\mathimg{D}_t^-$ w.r.t. the corresponding attributes.

\vspace{-2mm}
\section{Experiments}

\subsection{Settings}
\myparaheadd{Datasets.}
Our experiments are conducted on the MIT-Adobe FiveK dataset~\cite{bychkovsky2011learning} and the PPR10K dataset~\cite{liang2021ppr10k}.
The MIT-Adobe FiveK dataset contains 5,000 RAW images.
For each RAW image,
five reference images retouched by different experts (A/B/C/D/E) are provided.
We follow the pre-processing pipeline in~\cite{song2021starenhancer,wang2019underexposed},
and split the dataset into 4,500 pairs for training and 500 pairs for validation,
which is also known as MIT-Adobe-5K-UPE.
%
Both 340p (short side of the images) and full resolution are used for validation.
The PPR10K dataset~\cite{liang2021ppr10k} contains 11,161 portrait photos with 1,681 groups, and each photo has three retouched versions processed by three experts (A/B/C).
Following~\cite{liang2021ppr10k},
we divide the PPR10K dataset into a training set with 1,356 groups and 8,875 photos, and a testing set with 325 groups and 2,286 photos.
We employ the 360p setting in~\cite{liang2021ppr10k}.
For both datasets,
we construct the image-condition pairs for each image following the practice in \secref{sec:contrastive}.

\myparaheadd{Evaluation Metrics.}
To align with previous methods~\cite{song2021starenhancer,ouyang2023rsfnet},
we use the PSNR, SSIM, and LPIPS~\cite{zhang2018unreasonable} as the evaluation metrics.
In order to evaluate the similarity between the distribution of the results and that of expert-retouched GTs,
perceptual metric FID~\cite{heusel2017gans} is employed.
Note that the reference images for FID include all expert-retouched results.
We also adopt the no-reference metric NIMA~\cite{talebi2018nima} to evaluate the results from an aesthetic perspective.
All the metrics are implemented by IQA-PyTorch~\cite{pyiqa}.

\myparaheadd{Details for Training and Implementation.} 
Our \ourmethod~is built upon Stable Diffusion 2.1-base.
To accelerate the training process,
the pre-trained Stable Diffusion is adopted as the parameter initialization of our baseline model.
The training process lasts for approximately 200 epochs with a batch size of 64.
Following Stable Diffusion,
we use Adam~\cite{kingma2014adam} optimizer and the learning rate is set to $1\times 10^{-6}$.
The experiments are implemented on 8 NVIDIA Tesla 32G-V100 GPUs.
During training, 
the images are resized to $512\times512$ before being fed into the network.
After being encoded by the pre-trained encoder provided by Stable Diffusion,
the latent used for the training of the diffusion process is with the shape of $64\times 64\times 4$.
Following HDRNet~\cite{gharbi2017deep},
the shape of the affine bilateral grid is set to $16\times16\times8\times12$.
During inference, we adopt
the improved DDPM sampling~\cite{nichol2021improved} with 20 timesteps.
The input image $\mathimg{R}$ is resized to $64\times64$ to concatenate with the noisy latent $\mathimg{Z}$.
The obtained affine bilateral grid is sliced and applied to $\mathimg{R}$ of the original shape.
%
%
The hyper-parameters $\lambda$ and $\beta$ are set to 1 and 0.01, respectively.
The temperature parameter $\tau$ is set as 0.1.
\begin{table*}[tb]
\scriptsize
\centering
\newcommand{\kcl}{$^{\dag}$~}
\newcommand{\kcg}{$^{\S}$~}
\caption{
    Quantitative comparison on MIT-Adobe FiveK dataset with subsets retouched by five experts (A/B/C/D/E). Symbol \kcl represents the model trained with the mixture of five subsets. The best result is in \best{red} whereas the second is in \secondbest{blue}. The evaluations are done on the 340p setting.
}
\begin{tabular}{c|c|c|c|c|c|c|c|c}
    \toprule
    \multirow{2}{*}{Method} & \multicolumn{6}{c|}{PSNR$\uparrow$/SSIM$\uparrow$}  & \multirow{2}{*}{FID$\downarrow$} & \multirow{2}{*}{NIMA$\uparrow$} \\ \cline{2-7}
    & A    & B    & C   & D   & E   & Average                        &                      &                       \\
    \midrule
    PIENet\kcl~\cite{kim2020pienet}              & \secondbest{21.54}/\secondbest{0.882}        & \secondbest{26.02}/0.948       & 25.29/0.919       & 22.95/0.905       & 24.22/0.925       & 24.00/0.916       & 14.507       & 4.338           \\
    TSFlow\kcl~\cite{wang2022learning}               & 20.65/0.869        & 25.34/\secondbest{0.952}       & 25.57/0.935       & 22.48/0.913       & 23.65/0.935       & 23.54/0.921       & 9.678       & 4.976            \\
    StarEnhancer\kcl ~\cite{song2021starenhancer}& 20.75/0.880        & 25.84/\secondbest{0.952}       & \secondbest{25.73}/\secondbest{0.937}       & \secondbest{23.50}/\secondbest{0.922}       & \secondbest{24.60}/\secondbest{0.947}       & \secondbest{24.09}/\secondbest{0.928}       & \secondbest{9.493}       & \secondbest{4.977}           \\
    DiffRetouch\kcl                              & \best{24.48}/\best{0.936}        & \best{26.12}/\best{0.958}       & \best{26.21}/\best{0.944}       & \best{24.51}/\best{0.940}       & \best{24.67}/\best{0.953}       & \best{25.20}/\best{0.946}       & \best{8.957}       & \best{5.022}           \\
    \bottomrule
\end{tabular}
\label{tab:adobe_quantity2}
\vspace{-4mm}
\end{table*}

\setlength{\tabcolsep}{6pt}
\renewcommand{\arraystretch}{0.9}
\begin{table}[tb]
\centering
\newcommand{\kcl}{$^{\dag}$~}
\newcommand{\kcg}{$^{\S}$~}
\caption{
    Quantitative comparison on MIT-Adobe FiveK with Expert-C subset. \kcl represents the model trained with the mixture of five subsets. * represents these results are replicated from~\cite{song2021starenhancer, wang2022learning}. The best is in \best{red} whereas the second is in \secondbest{blue}. Both 340p (the shorter edge of the image) and full resolution are used for validation. The training codes of PIENet~\cite{kim2020pienet} is not yet accessible, and only the 340p results are released.}

\begin{tabular}{c|c|c|c|c|c|c}
    \toprule
    \multirow{2}{*}{Method} & \multicolumn{3}{c|}{340p} & \multicolumn{3}{c}{Full Resolution} \\
    \cline{2-7}
                        & PSNR$\uparrow$   & SSIM$\uparrow$   & LPIPS$\downarrow$    & PSNR $\uparrow$      & SSIM $\uparrow$    & LPIPS$\downarrow$       \\
    \midrule
    HDRNet~\cite{gharbi2017deep}            & 23.71    & 0.899    & 0.080    & 23.20*  & 0.917* & 0.120*            \\
    DeepUPE~\cite{wang2019underexposed}     & 23.48    & 0.907    & 0.085    & 23.24*  & 0.893* & 0.158*           \\
    CURL~\cite{moran2021curl}               & 24.40    & 0.935    & 0.061    & 24.20*  & 0.880* & 0.108*           \\
    DeepLPF~\cite{moran2020deeplpf}         & 24.43    & \secondbest{0.937}  & 0.059 & 24.48*  & 0.887* & 0.103*           \\
    3DLUT~\cite{zeng2020learning}           & 25.07    & \secondbest{0.937}    & 0.055    & 24.92*  & 0.934* & 0.093*           \\
    CSRNet~\cite{he2020conditional}         & 25.55    & 0.936    & 0.057    & 25.06*  & 0.935* & 0.090*           \\
    RSFNet~\cite{ouyang2023rsfnet}          & 25.34    & 0.938    & \best{0.051}    & 25.09 & 0.915 & \secondbest{0.081}            \\
    PIENet\kcl~\cite{kim2020pienet}         & 25.29    & 0.920    & 0.099    & -  & - & -            \\
    TSFlow\kcl~\cite{wang2022learning}          & 25.57    & 0.935    & 0.055    & \secondbest{25.36}*  & \secondbest{0.944}* & \best{0.079}*    \\
    StarEnhancer\kcl ~\cite{song2021starenhancer}& \secondbest{25.73}    & \secondbest{0.937}    & 0.055    & 25.29*  & 0.943* & 0.086*      \\
    DiffRetouch\kcl                         & \best{26.21}    & \best{0.944}    & \secondbest{0.054}    & \best{25.41}  & \best{0.952} & 0.088           \\
    \bottomrule
\end{tabular}
\vspace{-7mm}
\label{tab:adobe_quantity1}
\end{table}

\begin{figure}[th]
    \centering
    \includegraphics[width=0.98\linewidth]{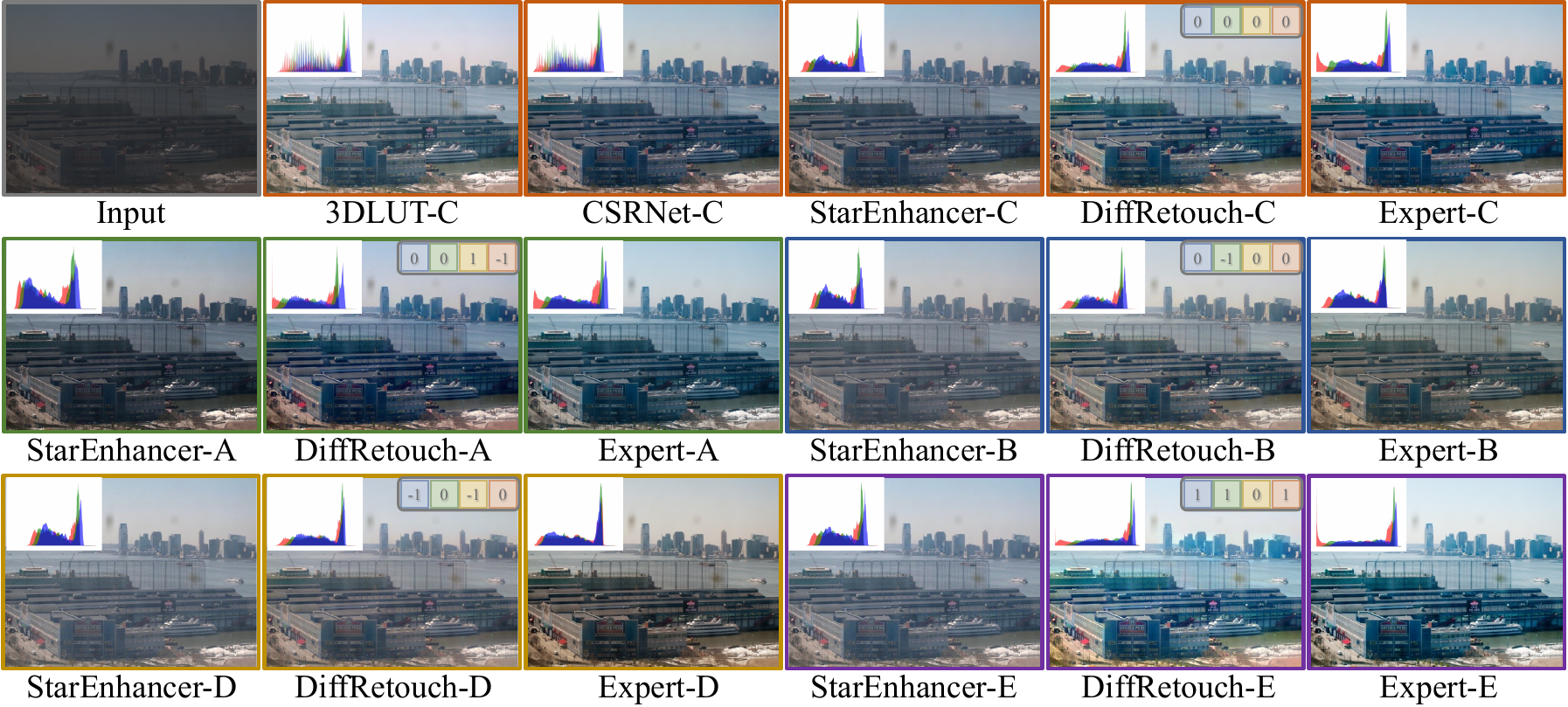}
    \caption{Qualitative comparison on MIT-Adobe FiveK dataset with subsets retouched by five experts (A/B/C/D/E). Since 3D-LUT~\cite{zeng2020learning} and CSRNet~\cite{he2020conditional} are unable to produce multiple retouching styles, only the results corresponding to Expert-C are displayed. The input condition $\mathvec{c}$ is shown at the top of each DiffRetouch generated result, along with the color histogram shown at the bottom left corner of the images. Results generated by our \ourmethod~are more similar to the corresponding expert-retouched result, especially for the color histogram.}
    \label{fig:adobe_quality}
    \vspace{-3mm}
\end{figure}

\begin{figure}[th]
    \centering
    \includegraphics[width=0.97\linewidth]{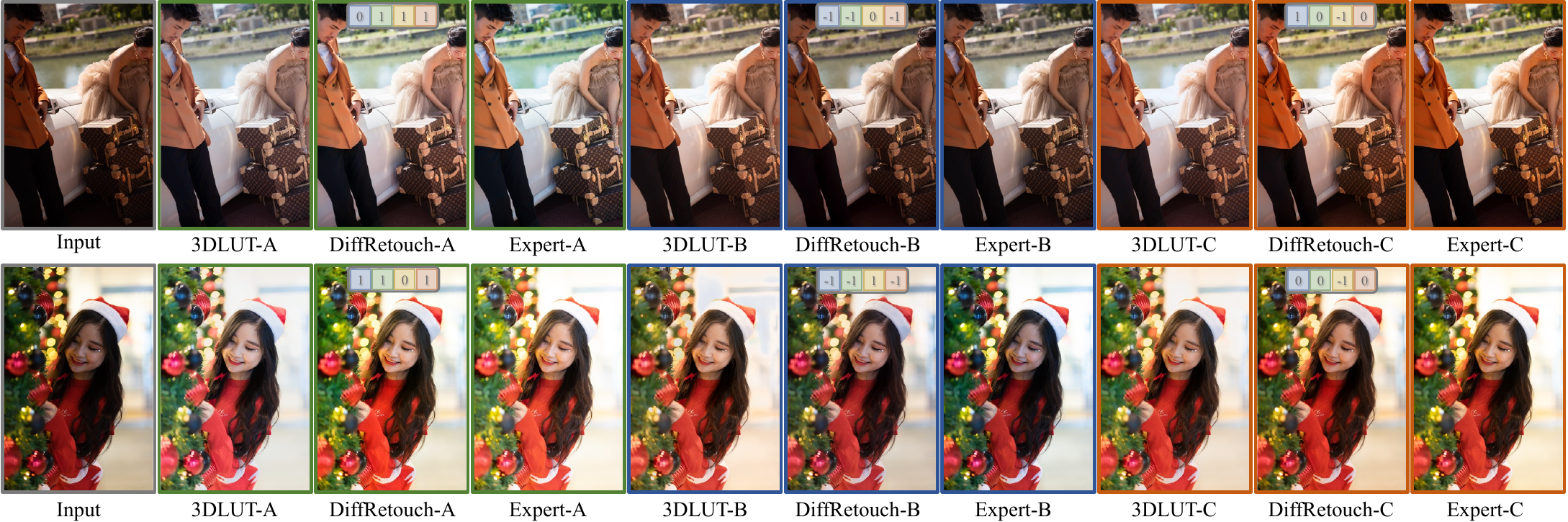}
    \caption{Qualitative comparison on PPR10K dataset with subsets retouched by three experts (A/B/C). The input condition $\mathvec{c}$ is shown at the top of each DiffRetouch generated result. Results generated by our \ourmethod~are more similar to the corresponding expert-retouched result.}
    \label{fig:ppr_quality}
    \vspace{-7mm}
\end{figure}

\renewcommand{\arraystretch}{0.9}
\begin{table}[th]
\centering
\newcommand{\kcl}{$^{\dag}$~}
\newcommand{\kcg}{$^{\S}$~}
\caption{
    Quantitative comparisons on PPR10K dataset with subsets retouched by three experts (A/B/C). The best result is in \best{red} whereas the second is in \secondbest{blue}. \kcg means evaluation with three models trained by three subsets respectively. \kcl means evaluation with one model trained with the mixture of three subsets.  Despite using only one model, our \ourmethod~achieves better or comparable performance against other methods with three models.
}
\begin{tabular}{ccccccc}
\toprule
Method            & Dataset & PSNR$\uparrow$ & SSIM$\uparrow$ & LPIPS$\downarrow$ & FID$\downarrow$ & NIMA$\uparrow$               \\
\midrule
\multirow{3}{*}{HDRNet\kcg~\cite{gharbi2017deep}} & Expert-A      & 23.01     & 0.953     & 0.057      & \multirow{3}{*}{2.782}  & \multirow{3}{*}{5.490} \\
                  & Expert-B        & 23.17     & 0.952     & 0.058      &       &            \\
                  & Expert-C        & 23.34     & 0.951     & 0.058      &       &            \\
\midrule
\multirow{3}{*}{CSRNet\kcg~\cite{he2020conditional}} & Expert-A        & 23.86     & 0.952     & 0.055      & \multirow{3}{*}{3.077}      & \multirow{3}{*}{5.465} \\
                  & Expert-B        & 23.70     & 0.952     & 0.057      &                   \\
                  & Expert-C        & 23.87     & 0.953     & 0.055      &                   \\
\midrule
\multirow{3}{*}{RSFNet\kcg~\cite{ouyang2023rsfnet}} & Expert-A        & 25.58     & 0.965     & 0.043      & \multirow{3}{*}{\best{2.467}}      & \multirow{3}{*}{\best{5.516}} \\
                  & Expert-B        & 24.81     & \secondbest{0.960}     & 0.048      &                   \\
                  & Expert-C        & 25.56     & \secondbest{0.964}     & \secondbest{0.044}      &                   \\
\midrule
\multirow{3}{*}{3DLUT\kcg~\cite{zeng2020learning}} & Expert-A        & \secondbest{25.98}     & \secondbest{0.967}     & \secondbest{0.040}      & \multirow{3}{*}{2.485}      & \multirow{3}{*}{5.492} \\
                  & Expert-B        & \secondbest{25.06}     & 0.959     & \secondbest{0.046}      &       &            \\
                  & Expert-C        & \secondbest{25.45}     & 0.961     & 0.045     &       &            \\
\midrule
\multirow{3}{*}{DiffRetouch\kcl} & Expert-A        & \best{26.26}     & \best{0.972}     & \best{0.038}      & \multirow{3}{*}{\secondbest{2.477}}      & \multirow{3}{*}{\secondbest{5.514}} \\
                  & Expert-B        & \best{25.52}     & \best{0.969}     & \best{0.040}      &         &          \\
                  & Expert-C        & \best{25.76}     & \best{0.967}     & \best{0.039}     &         &          \\  
\bottomrule
\end{tabular}
\vspace{-4mm}
\label{tab:ppr_quantity}
\end{table}


\subsection{Comparison with Other Methods}
We compare our \ourmethod~with other methods proposed for retouching.
Based on the ability to generate diverse retouched results,
we classify these methods into two categories.
The first type of method adopts deterministic models,
and can only produce a single retouching style for each input,
which includes HDRNet~\cite{gharbi2017deep},
DeepUPE~\cite{wang2019underexposed},
CURL~\cite{moran2021curl},
DeepLPF~\cite{moran2020deeplpf},
3DLUT~\cite{zeng2020learning},
CSRNet~\cite{he2020conditional},
and RSFNet~\cite{ouyang2023rsfnet}.
Following the default setting,
these methods train multiple models for subsets retouched by different experts, and evaluate separately on each subset.
The other type of methods,
like PIENet~\cite{kim2020pienet}, StarEnhancer~\cite{song2021starenhancer}, and TSFlow~\cite{wang2022learning},
supports generating multiple retouching styles.
%
These methods adopt one model trained with results retouched by different experts.
%
During inference, to generate the corresponding predictions for each expert,
style embedding~\cite{song2021starenhancer,wang2022learning}, which represents the overall retouching style of the expert, is extracted from additional images and sent into their models.
Thanks to the superiority of our editing mechanism,
we can generate the desired retouching style by adjusting condition $\mathvec{c}$,
without the need for additional images.
To evaluate the multi-style retouching performance of our \ourmethod,
we provide the model with the pre-calculated condition of GT styles,
thus making it produce corresponding output for evaluation.
For each input image,
we follow the practice in \secref{sec:contrastive} to construct $\mathvec{c}$ for different experts.
%
%
Briefly,
we first calculate their $\mathvec{s}_i$ for each expert-retouched result.
Then for each attribute ($i$),
the corresponding coefficient $\mathvec{c}_i$ is set to 1(-1) for the GT with the highest(lowest) $\mathvec{s}_i$.
Note that when used in practice, these conditions are provided and adjusted by the users according to their aesthetic preferences.
%
%


For MIT-Adobe FiveK,
the qualitative and quantitative comparisons are shown in \figref{fig:adobe_quality}, \tabref{tab:adobe_quantity2}, and \tabref{tab:adobe_quantity1}.
\tabref{tab:adobe_quantity1} shows the comparison against the state-of-the-art methods evaluated on the Expert-C subset.
For methods that support various styles,
the performance of generating retouching styles of five experts are shown in \tabref{tab:adobe_quantity2}.
%
%
We also compare our \ourmethod~with HDRNet~\cite{gharbi2017deep}, CSRNet~\cite{he2020conditional}, RSFNet~\cite{ouyang2023rsfnet}, and 3DLUT~\cite{zeng2020learning} on PPR10K.
Note that for subsets of three experts(A/B/C),
these methods are evaluated with three models trained on three subsets respectively,
while our \ourmethod~is a single model trained with a mixture of subsets.
As can be seen in \tabref{tab:ppr_quantity},
despite using only one model against three models of other methods,
our method achieves better or comparable performance on referenced and non-referenced evaluation metrics.
The visual results on both datasets are shown in \figref{fig:adobe_quality} and \figref{fig:ppr_quality}.

\begin{figure}[t]
    \centering
    \includegraphics[width=\linewidth]{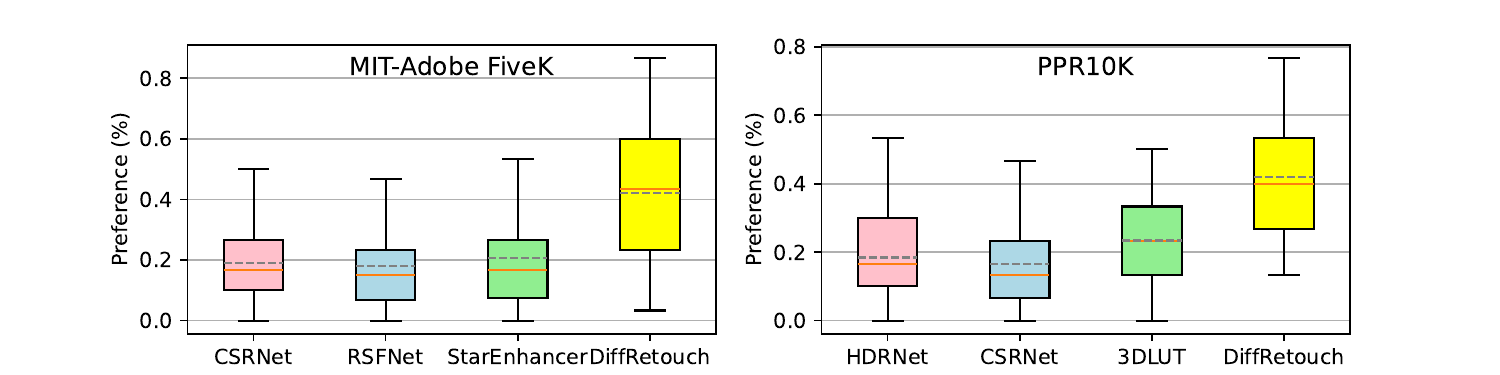}
    \caption{Boxplot of our user study. The dashed gray line and the solid red line inside the box are the mean and the median preference percentages respectively.}
    \label{fig:user}
    \vspace{-7mm}
\end{figure}

To further validate the effectiveness of our method,
we conduct a user study to evaluate human preferences for our \ourmethod~and other state-of-the-art methods,
including CSRNet~\cite{he2020conditional}, RSFNet~\cite{ouyang2023rsfnet}, StarEnhancer~\cite{song2021starenhancer}, HDRNet~\cite{gharbi2017deep}, 3DLUT~\cite{zeng2020learning}.
We randomly select 30 images each from  MIT-Adobe FiveK and PPR10K datasets for a total of 60 images,
and invite 50 volunteers to participate in the user study.
Given the original images and retouched results generated by all methods,
these participants are required to choose their preferred result.
We calculate the preference percentages of participants for each method and plot the results shown in \figref{fig:user}.
Among these methods,
our \ourmethod~achieves the highest mean and median percentages, 
almost twice that of the second place,
which demonstrates the superior performance of our method.

\vspace{-4mm}
\subsection{Ablation Study}
As shown in \tabref{tab:ablation1},
we carry out the ablation study to demonstrate the necessity of each design adopted in \ourmethod.
Three metrics are chosen to evaluate the results on both Adobe5K and PPR10K datasets.

\begin{figure}[tb]
  \centering
  \begin{minipage}{0.55\linewidth}
    \includegraphics[width=\linewidth]{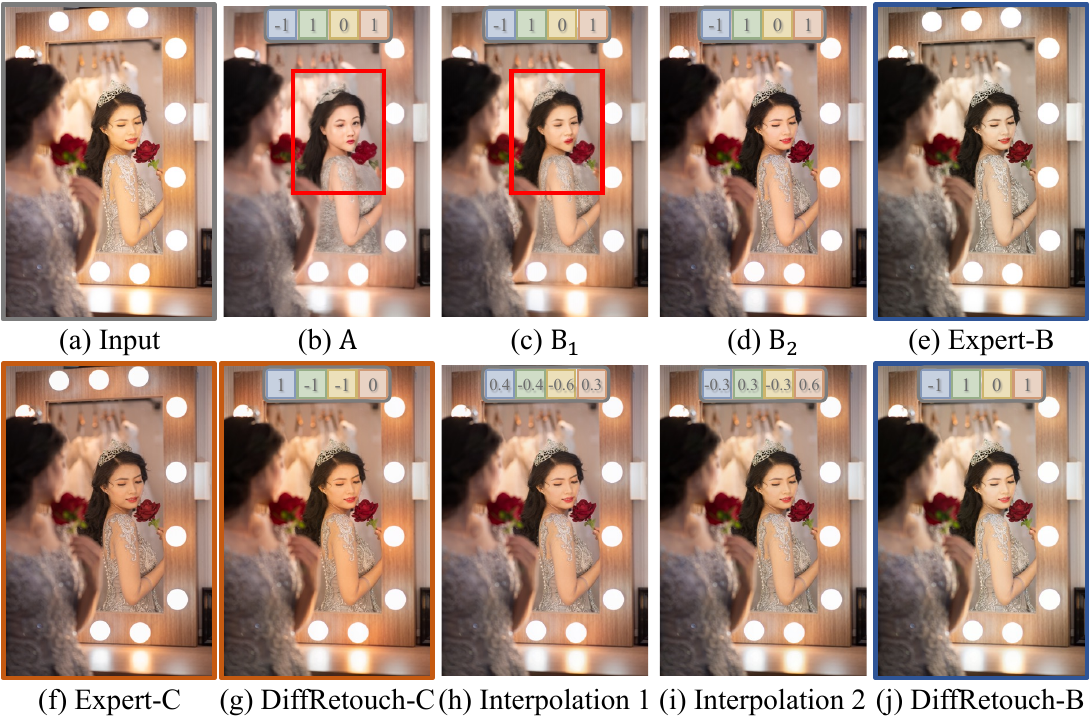}
    \caption{Ablation study of our model. `$\rm A$' represents the baseline model. `$\rm B_1$' and `$\rm B_2$' represent the baseline model equipped with CFW~\cite{wang2023exploiting} and affine bilateral grid, but both w/o $\mathcal{L}_{cl}$. (g)-(j) show the intermediate retouching styles generated by our full model w/ $\mathcal{L}_{cl}$ through interpolating the coefficients.}
    \label{fig:ablation}
  \end{minipage}
  \hfill
  \begin{minipage}{0.41\linewidth}
    \includegraphics[width=\linewidth]{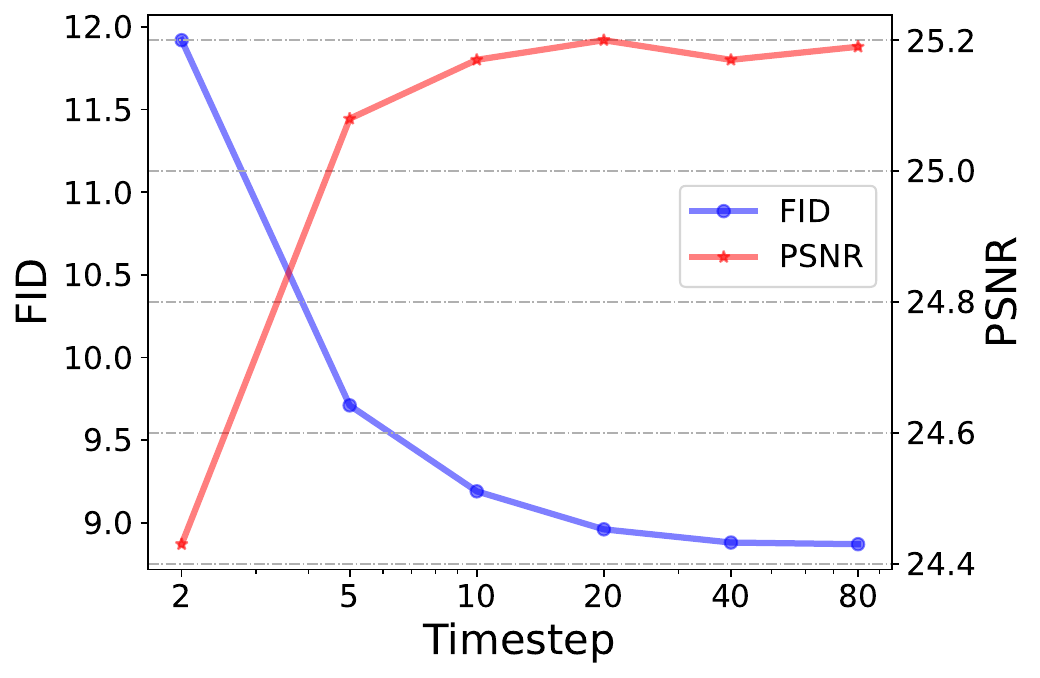}
    \caption{Effect of the sampling timesteps. The blue and red lines are the values of FID and PSNR, respectively. The performance of \ourmethod~significantly improves before the timesteps increase to 20. To balance efficiency and effectiveness, the default number of sampling timestep is set as 20 in this paper.
   }
    \label{fig:step}
  \end{minipage}
  \vspace{-3mm}
\end{figure}

\setlength{\tabcolsep}{6pt}
\renewcommand{\arraystretch}{0.9}
\begin{table}[t]
\centering
\caption{Quantitative results of our model and its variants on Adobe5K and PPR10K datasets. `$\rm A$' represents the baseline model. `$\rm B_1$' and `$\rm B_2$' represent the baseline model equipped with CFW~\cite{wang2023exploiting} and affine bilateral grid respectively, but both w/o $\mathcal{L}_{cl}$. `$\rm C$' represents the full model with the contrastive learning scheme.}
\begin{tabular}{c|c|c|c|c|c|c}
\toprule
\multirow{2}{*}{Model} & \multicolumn{3}{c|}{Adobe5K} & \multicolumn{3}{c}{PPR10K} \\ \cline{2-7}
                       & PSNR$\uparrow$  & SSIM$\uparrow$   & FID$\downarrow$ & PSNR$\uparrow$  & SSIM$\uparrow$  & FID$\downarrow$ \\
\midrule
$\rm A$              & 19.83      & 0.618      & 50.157    & 20.47      & 0.694         & 16.673    \\
$\rm B_1$            & 20.97      & 0.885      & 26.721    & 21.79      & 0.831         & 14.511    \\
$\rm B_2$            & 23.70      & 0.936      & 11.476    & 25.15      & 0.964         & 3.091    \\
$\rm C$              & 25.20      & 0.946      & 8.957     & 25.85      & 0.969         & 2.499    \\
\bottomrule
\end{tabular}
\vspace{-6mm}
\label{tab:ablation1}
\end{table}

\setlength{\tabcolsep}{6pt}
\renewcommand{\arraystretch}{0.9}
\begin{table}[tb]
\centering
\caption{
    Comparison of \textbf{adjustable range} w/ and w/o contrastive learning scheme. `$\rm B_2$' represents the baseline model equipped with the affine bilateral grid.  `$\rm C$' represents the full model with the contrastive learning scheme. The adjustable range for each attribute ($i$) is determined by the changes in the corresponding scores ($\mathvec{s}_i$) of results when varying the coefficient ($\mathvec{c}_i$) from maximum (1) to minimum (-1). A larger adjustable range means more significant influence brought by adjusting the coefficient, proving the effectiveness of the contrastive learning scheme.
}
\newcommand{\kcl}{$^{\dag}$~}
\newcommand{\kcg}{$^{\S}$~}
\begin{tabular}{c|c|c|c|c}
    \toprule
    Model    & Colorfulness$\uparrow$           & Contrast$\uparrow$      & \makecell{ Color \\ Temperature}$\uparrow$     & Brightness$\uparrow$     \\
    \midrule
    $\rm B_2$  &   10.12            &   252.7              &   1141.5              &  26.46      \\
    $\rm C$    &   20.60           &    297.3              &   1572.9              &  35.22       \\      
    \bottomrule
\end{tabular}
\label{tab:ablation2}
\vspace{-7mm}
\end{table}

\myparaheadd{Introduction of affine bilateral grid.} 
Our intention of introducing the affine bilateral grid is to solve the texture distortion existing in our baseline model.
Motivated by StableSR~\cite{wang2023exploiting},
the controllable feature wrapping (CFW) module is adopted,
which extracts the intermediate features from the encoder to modulate the decoder features.
As shown in \figref{fig:ablation} (c),
the severe texture distortion can be somewhat alleviated.
However, it cannot handle complex structures like human faces.
With the introduction of the affine bilateral grid,
this problem can be well solved and original details are reserved,
which is shown in \figref{fig:ablation} (d).
%
The improvements in \tabref{tab:ablation1} on both datasets are significant,
which demonstrates the effectiveness of the affine bilateral grid.

\myparaheadd{Contrastive learning scheme.}
As shown in \figref{fig:ablation} (d),
%
the influence brought by adjusting the attributes is too weak to reach the retouching style of experts in \figref{fig:ablation} (e).
With the implementation of a contrastive learning scheme,
our \ourmethod~is encouraged to be more aware of the adjustment brought by each attribute and outputs results more like experts especially for the color temperature,
which is shown in \figref{fig:ablation} (j).
%
To better show the effectiveness of contrastive learning,
we attempt to quantify the influence brought by adjusting each attribute.
For each attribute ($i$), we set the corresponding $\mathvec{c}_i$ to the maximum (1) and minimum (-1) values with other three remaining 0,
and generate two extreme results.
By calculating the difference in corresponding scores ($\mathvec{s}_i$) of these two results,
the adjustable range \wrt this attribute can be obtained.
The average adjustable ranges for Adobe5K before and after the implementation of contrastive learning are shown in \tabref{tab:ablation2},
and we can see that the adjustable ranges are enlarged for four attributes.
The improvements of FID for model $\rm B_2$ and $\rm C$ in \tabref{tab:ablation1} further prove that the contrastive learning helps the model better fit the expert-retouched distribution.

\myparaheadd{Adjustable condition.}
Our editing mechanism allows users to customize their preferred style by adjusting the coefficients \wrt four attributes.
\figref{fig:ablation} (g)-(j) show an example of generating the intermediate styles between the style of Expert B and C.
%
As the coefficients are adjusted.
we can see that the style changes continuously and in the correct direction for the corresponding attribute.

%
%


\myparaheadd{Sampling timesteps.}
During inference, 
we adopt the improved DDPM~\cite{nichol2021improved}, 
which can reduce the original timesteps.
In order to investigate the effect of the sampling timestep,
we set different sampling steps and evaluate the results,
and the results are shown in~\figref{fig:step}.
We can see that the performance of \ourmethod~significantly improves as the timesteps increase to 20.
As the number of timesteps continues increasing above 20 timesteps,
the improvements become limited.
To balance efficiency and effectiveness,
the default number of sampling timestep is set as 20 in this paper.
%
%
%
We also test the inference speed on a single Tesla 32G-V100 GPU,
and it takes about 1 second when adopting 20 timesteps for a $1920 \times 1080$ image,
which is enough to meet the needs of practical applications.
For most other retouching methods like CSRNet~\cite{he2020conditional} and RSFNet~\cite{ouyang2023rsfnet}, the inference time will increase when handling images with higher resolution.
%
However, our DiffRetouch can be easily applied to images of any resolution with the operation of slicing~\cite{gharbi2017deep}.
Moreover, since the input image is resized to a fixed $64 \times 64$ resolution before being sent to the denoising U-Net and the size of the affine bilateral grid is also fixed,
the inference time is marginally affected by the input size.

\vspace{-4mm}
\section{Conclusion}
In this paper,
we propose a diffusion-based method for image retouching,
called \ourmethod.
Considering the subjectivity of this task,
we leverage the excellent distribution coverage of diffusion to capture the fine-retouched distribution,
allowing to sample various visual-pleasing styles.
Four adjustable coefficients \wrt four image attributes are provided for users to edit the final results.
The affine bilateral grid and contrastive learning scheme are introduced to address the texture distortion and control insensitivity.
Extensive experiments have demonstrated the superiority of our \ourmethod~and the effectiveness of each component.

%
%
\bibliographystyle{splncs04}
\bibliography{main}

\begin{thebibliography}{10}
\providecommand{\url}[1]{\texttt{#1}}
\providecommand{\urlprefix}{URL }
\providecommand{\doi}[1]{https://doi.org/#1}

\bibitem{bychkovsky2011learning}
Bychkovsky, V., Paris, S., Chan, E., Durand, F.: Learning photographic global tonal adjustment with a database of input/output image pairs. In: CVPR (2011)

\bibitem{pyiqa}
Chen, C., Mo, J.: {IQA-PyTorch}: Pytorch toolbox for image quality assessment. [Online]. Available: \url{https://github.com/chaofengc/IQA-PyTorch} (2022)

\bibitem{chen2018learning}
Chen, C., Chen, Q., Xu, J., Koltun, V.: Learning to see in the dark. In: CVPR (2018)

\bibitem{chen2018deep}
Chen, Y.S., Wang, Y.C., Kao, M.H., Chuang, Y.Y.: Deep photo enhancer: Unpaired learning for image enhancement from photographs with gans. In: CVPR (2018)

\bibitem{chen2023image}
Chen, Z., Zhang, Y., Gu, J., Yuan, X., Kong, L., Chen, G., Yang, X.: Image super-resolution with text prompt diffusion. arXiv preprint arXiv:2311.14282  (2023)

\bibitem{chung2022improving}
Chung, H., Sim, B., Ryu, D., Ye, J.C.: Improving diffusion models for inverse problems using manifold constraints. In: NeurIPS (2022)

\bibitem{fei2023generative}
Fei, B., Lyu, Z., Pan, L., Zhang, J., Yang, W., Luo, T., Zhang, B., Dai, B.: Generative diffusion prior for unified image restoration and enhancement. In: CVPR (2023)

\bibitem{gharbi2017deep}
Gharbi, M., Chen, J., Barron, J.T., Hasinoff, S.W., Durand, F.: Deep bilateral learning for real-time image enhancement. ACM TOG  (2017)

\bibitem{guo2023shadowdiffusion}
Guo, L., Wang, C., Yang, W., Huang, S., Wang, Y., Pfister, H., Wen, B.: Shadowdiffusion: When degradation prior meets diffusion model for shadow removal. In: CVPR (2023)

\bibitem{colorfulness}
Hasler, D., Suesstrunk, S.E.: Measuring colorfulness in natural images. In: Human vision and electronic imaging VIII (2003)

\bibitem{he2020conditional}
He, J., Liu, Y., Qiao, Y., Dong, C.: Conditional sequential modulation for efficient global image retouching. In: ECCV (2020)

\bibitem{hernandez1999calculating}
Hernandez-Andres, J., Lee, R.L., Romero, J.: Calculating correlated color temperatures across the entire gamut of daylight and skylight chromaticities. Applied optics  (1999)

\bibitem{heusel2017gans}
Heusel, M., Ramsauer, H., Unterthiner, T., Nessler, B., Hochreiter, S.: Gans trained by a two time-scale update rule converge to a local nash equilibrium. In: NeurIPS (2017)

\bibitem{ho2020denoising}
Ho, J., Jain, A., Abbeel, P.: Denoising diffusion probabilistic models. In: NeurIPS (2020)

\bibitem{hou2023global}
Hou, J., Zhu, Z., Hou, J., Liu, H., Zeng, H., Yuan, H.: Global structure-aware diffusion process for low-light image enhancement. arXiv preprint arXiv:2310.17577  (2023)

\bibitem{hu2018exposure}
Hu, Y., He, H., Xu, C., Wang, B., Lin, S.: Exposure: A white-box photo post-processing framework. ACM TOG  (2018)

\bibitem{jiang2023autodir}
Jiang, Y., Zhang, Z., Xue, T., Gu, J.: Autodir: Automatic all-in-one image restoration with latent diffusion. arXiv preprint arXiv:2310.10123  (2023)

\bibitem{kawar2022denoising}
Kawar, B., Elad, M., Ermon, S., Song, J.: Denoising diffusion restoration models. In: NeurIPS (2022)

\bibitem{kim2020pienet}
Kim, H.U., Koh, Y.J., Kim, C.S.: Pienet: Personalized image enhancement network. In: ECCV (2020)

\bibitem{kim2021representative}
Kim, H., Choi, S.M., Kim, C.S., Koh, Y.J.: Representative color transform for image enhancement. In: ICCV (2021)

\bibitem{kim2023learning}
Kim, H., Lee, K.M.: Learning controllable isp for image enhancement. IEEE TIP  (2023)

\bibitem{kingma2014adam}
Kingma, D.P., Ba, J.: Adam: A method for stochastic optimization. arXiv preprint arXiv:1412.6980  (2014)

\bibitem{kosugi2020unpaired}
Kosugi, S., Yamasaki, T.: Unpaired image enhancement featuring reinforcement-learning-controlled image editing software. In: AAAI (2020)

\bibitem{li2020flexible}
Li, C., Guo, C., Ai, Q., Zhou, S., Loy, C.C.: Flexible piecewise curves estimation for photo enhancement. arXiv preprint arXiv:2010.13412  (2020)

\bibitem{li2022cudi}
Li, C., Guo, C., Feng, R., Zhou, S., Loy, C.C.: Cudi: Curve distillation for efficient and controllable exposure adjustment. arXiv preprint arXiv:2207.14273  (2022)

\bibitem{li2021learning}
Li, C., Guo, C., Loy, C.C.: Learning to enhance low-light image via zero-reference deep curve estimation. IEEE TPAMI  (2021)

\bibitem{liang2021ppr10k}
Liang, J., Zeng, H., Cui, M., Xie, X., Zhang, L.: Ppr10k: A large-scale portrait photo retouching dataset with human-region mask and group-level consistency. In: CVPR (2021)

\bibitem{lin2023diffbir}
Lin, X., He, J., Chen, Z., Lyu, Z., Fei, B., Dai, B., Ouyang, W., Qiao, Y., Dong, C.: Diffbir: Towards blind image restoration with generative diffusion prior. arXiv preprint arXiv:2308.15070  (2023)

\bibitem{liu2021retinex}
Liu, R., Ma, L., Zhang, J., Fan, X., Luo, Z.: Retinex-inspired unrolling with cooperative prior architecture search for low-light image enhancement. In: CVPR (2021)

\bibitem{lugmayr2022repaint}
Lugmayr, A., Danelljan, M., Romero, A., Yu, F., Timofte, R., Van~Gool, L.: Repaint: Inpainting using denoising diffusion probabilistic models. In: CVPR (2022)

\bibitem{luo2023image}
Luo, Z., Gustafsson, F.K., Zhao, Z., Sj{\"o}lund, J., Sch{\"o}n, T.B.: Image restoration with mean-reverting stochastic differential equations. arXiv preprint arXiv:2301.11699  (2023)

\bibitem{van2008visualizing}
Van~der Maaten, L., Hinton, G.: Visualizing data using t-sne. Journal of machine learning research  (2008)

\bibitem{meng2022diffusion}
Meng, X., Kabashima, Y.: Diffusion model based posterior sampling for noisy linear inverse problems. arXiv preprint arXiv:2211.12343  (2022)

\bibitem{moran2020deeplpf}
Moran, S., Marza, P., McDonagh, S., Parisot, S., Slabaugh, G.: Deeplpf: Deep local parametric filters for image enhancement. In: CVPR (2020)

\bibitem{moran2021curl}
Moran, S., McDonagh, S., Slabaugh, G.: Curl: Neural curve layers for global image enhancement. In: ICPR (2021)

\bibitem{nichol2021improved}
Nichol, A.Q., Dhariwal, P.: Improved denoising diffusion probabilistic models. In: ICML (2021)

\bibitem{oord2018representation}
Oord, A.v.d., Li, Y., Vinyals, O.: Representation learning with contrastive predictive coding. arXiv preprint arXiv:1807.03748  (2018)

\bibitem{ouyang2023rsfnet}
Ouyang, W., Dong, Y., Kang, X., Ren, P., Xu, X., Xie, X.: Rsfnet: A white-box image retouching approach using region-specific color filters. In: ICCV (2023)

\bibitem{ozdenizci2023restoring}
{\"O}zdenizci, O., Legenstein, R.: Restoring vision in adverse weather conditions with patch-based denoising diffusion models. IEEE TPAMI  (2023)

\bibitem{ren2023multiscale}
Ren, M., Delbracio, M., Talebi, H., Gerig, G., Milanfar, P.: Multiscale structure guided diffusion for image deblurring. In: ICCV (2023)

\bibitem{rombach2022high}
Rombach, R., Blattmann, A., Lorenz, D., Esser, P., Ommer, B.: High-resolution image synthesis with latent diffusion models. In: CVPR (2022)

\bibitem{saharia2022palette}
Saharia, C., Chan, W., Chang, H., Lee, C., Ho, J., Salimans, T., Fleet, D., Norouzi, M.: Palette: Image-to-image diffusion models. In: ACM SIGGRAPH (2022)

\bibitem{saharia2022image}
Saharia, C., Ho, J., Chan, W., Salimans, T., Fleet, D.J., Norouzi, M.: Image super-resolution via iterative refinement. IEEE TPAMI  (2022)

\bibitem{song2021starenhancer}
Song, Y., Qian, H., Du, X.: Starenhancer: Learning real-time and style-aware image enhancement. In: ICCV (2021)

\bibitem{sun2023coser}
Sun, H., Li, W., Liu, J., Chen, H., Pei, R., Zou, X., Yan, Y., Yang, Y.: Coser: Bridging image and language for cognitive super-resolution. arXiv preprint arXiv:2311.16512  (2023)

\bibitem{sun2021enhance}
Sun, X., Li, M., He, T., Fan, L.: Enhance images as you like with unpaired learning. arXiv preprint arXiv:2110.01161  (2021)

\bibitem{talebi2018nima}
Talebi, H., Milanfar, P.: Nima: Neural image assessment. IEEE TIP  (2018)

\bibitem{Tseng2022NeuralPhotoFinishing}
Tseng, E., Zhang, Y., Jebe, L., Zhang, C., Xia, Z., Fan, Y., Heide, F., Chen, J.: Neural photo-finishing. ACM TOG  (2022)

\bibitem{wang2022learning}
Wang, H., Zhang, J., Liu, M., Wu, X., Zuo, W.: Learning diverse tone styles for image retouching. arXiv preprint arXiv:2207.05430  (2022)

\bibitem{wang2023exploring}
Wang, J., Chan, K.C., Loy, C.C.: Exploring clip for assessing the look and feel of images. In: AAAI (2023)

\bibitem{wang2023exploiting}
Wang, J., Yue, Z., Zhou, S., Chan, K.C., Loy, C.C.: Exploiting diffusion prior for real-world image super-resolution. arXiv preprint arXiv:2305.07015  (2023)

\bibitem{wang2019underexposed}
Wang, R., Zhang, Q., Fu, C.W., Shen, X., Zheng, W.S., Jia, J.: Underexposed photo enhancement using deep illumination estimation. In: CVPR (2019)

\bibitem{wang2021real}
Wang, T., Li, Y., Peng, J., Ma, Y., Wang, X., Song, F., Yan, Y.: Real-time image enhancer via learnable spatial-aware 3d lookup tables. In: ICCV (2021)

\bibitem{wang2022zero}
Wang, Y., Yu, J., Zhang, J.: Zero-shot image restoration using denoising diffusion null-space model. arXiv preprint arXiv:2212.00490  (2022)

\bibitem{wang2023exposurediffusion}
Wang, Y., Yu, Y., Yang, W., Guo, L., Chau, L.P., Kot, A.C., Wen, B.: Exposurediffusion: Learning to expose for low-light image enhancement. In: ICCV (2023)

\bibitem{whang2022deblurring}
Whang, J., Delbracio, M., Talebi, H., Saharia, C., Dimakis, A.G., Milanfar, P.: Deblurring via stochastic refinement. In: CVPR (2022)

\bibitem{wu2023seesr}
Wu, R., Yang, T., Sun, L., Zhang, Z., Li, S., Zhang, L.: Seesr: Towards semantics-aware real-world image super-resolution. arXiv preprint arXiv:2311.16518  (2023)

\bibitem{xie2023smartbrush}
Xie, S., Zhang, Z., Lin, Z., Hinz, T., Zhang, K.: Smartbrush: Text and shape guided object inpainting with diffusion model. In: CVPR (2023)

\bibitem{yang2022adaint}
Yang, C., Jin, M., Jia, X., Xu, Y., Chen, Y.: Adaint: Learning adaptive intervals for 3d lookup tables on real-time image enhancement. In: CVPR (2022)

\bibitem{yang2023pixel}
Yang, T., Ren, P., Xie, X., Zhang, L.: Pixel-aware stable diffusion for realistic image super-resolution and personalized stylization. arXiv preprint arXiv:2308.14469  (2023)

\bibitem{yin2023cle}
Yin, Y., Xu, D., Tan, C., Liu, P., Zhao, Y., Wei, Y.: Cle diffusion: Controllable light enhancement diffusion model. In: ACM MM (2023)

\bibitem{yu2024scaling}
Yu, F., Gu, J., Li, Z., Hu, J., Kong, X., Wang, X., He, J., Qiao, Y., Dong, C.: Scaling up to excellence: Practicing model scaling for photo-realistic image restoration in the wild. arXiv preprint arXiv:2401.13627  (2024)

\bibitem{zeng2020learning}
Zeng, H., Cai, J., Li, L., Cao, Z., Zhang, L.: Learning image-adaptive 3d lookup tables for high performance photo enhancement in real-time. IEEE TPAMI  (2020)

\bibitem{zhang2018unreasonable}
Zhang, R., Isola, P., Efros, A.A., Shechtman, E., Wang, O.: The unreasonable effectiveness of deep features as a perceptual metric. In: CVPR (2018)

\bibitem{zhu2020zero}
Zhu, A., Zhang, L., Shen, Y., Ma, Y., Zhao, S., Zhou, Y.: Zero-shot restoration of underexposed images via robust retinex decomposition. In: ICME (2020)

\end{thebibliography}

\clearpage
\section*{\Large{Appendices}}
\appendix

\section{Details about Network Architecture}
\label{sec:net}
The overview of our \ourmethod~can be seen in Figure 2 in the main paper,
and here we provide more details about the network architecture.
Our \ourmethod~is built upon Stable Diffusion 2.1-base,
where the underlying backbone U-Net is almost the same as Stable Diffusion,
except that the numbers of the input channels and the output channels have been modified.
In our implementation,
there are two main condition inputs, where one is the adjustable vector $\mathvec{c}$ consisting of four coefficients \wrt four image attributes, while the other is the input image $\mathimg{R}$.
As mentioned in the main paper,
the condition input $\mathvec{c} = [\mathvec{c}_1,\mathvec{c}_2,\mathvec{c}_3,\mathvec{c}_4]$ is mapped to the intermediate layers of the backbone U-Net via a cross-attention mechanism, 
which can be formulated as $\text{Attention}(\mathimg{Q}, \mathimg{K}, \mathimg{V}) = \text{softmax}(\frac{\mathimg{Q}\mathimg{K}^T}{\sqrt{d}})\mathimg{V}$,
and
\begin{equation}
    \mathimg{Q} = \phi^{(i)} \mathimg{W_Q}^{(i)}, \mathimg{K} = \mathvec{c}\mathimg{W_K}^{(i)}, \mathimg{V} = \mathvec{c}\mathimg{W_V}^{(i)},
\end{equation}
where $\phi^{(i)} \in \mathbb{R}^{N\times d_{\bm{\epsilon}_\theta}^{(i)}}$ is the flattened intermediate representation of $i$-th layer of the denoising model $\bm{\epsilon}_\theta$,
and $N$ is the channel number.
$\mathimg{W_Q}^{(i)} \in \mathbb{R}^{d_{\bm{\epsilon}}^{(i)} \times d}$, $\mathimg{W_K}^{(i)} \in \mathbb{R}^{d_\mathvec{c} \times d}$, $\mathimg{W_V}^{(i)} \in \mathbb{R}^{d_\mathvec{c} \times d}$ are learnable projection matrices.
Note that all layers in $\bm{\epsilon}_\theta$ are implemented with the cross-attention mechanism.
$d_{\bm{\epsilon}}^{(i)}$ and $d_\mathvec{c}$ represent the dimension of the flattened $\bm{\epsilon}^{(i)}$ and $\mathvec{c}$, 
and $d$ denotes the dimension of our attention calculation.

Another condition is the input image $\mathimg{R}$,
which is resized and then fed into the underlying U-Net via concatenation.
Besides the noise estimation, the underlying U-Net outputs the affine bilateral grid as well.
Thus the input channel and the output channel of the U-Net have been modified.
Specifically,
the input channel is changed from 4 to 7, adding 3 channels of the resized image.
The output channel is changed from 4 to 8,
where the first 4 channels are the noise prediction, 
and the last 4 channels are $\tilde{\mathimg{A}}$.
%
After sending $\tilde{\mathimg{A}}$ into four convolutional layers,
where two are $3 \times 3$ convolutions followed by ReLU and two are $1 \times 1$ convolutions,
we can obtain the affine bilateral grid $\mathimg{A}$.
Then the details about the pixel-level network and how to utilize the affine bilateral grid are described in~\secref{sec:abg}.

\begin{figure*}
    \centering
    \includegraphics[width=\linewidth]{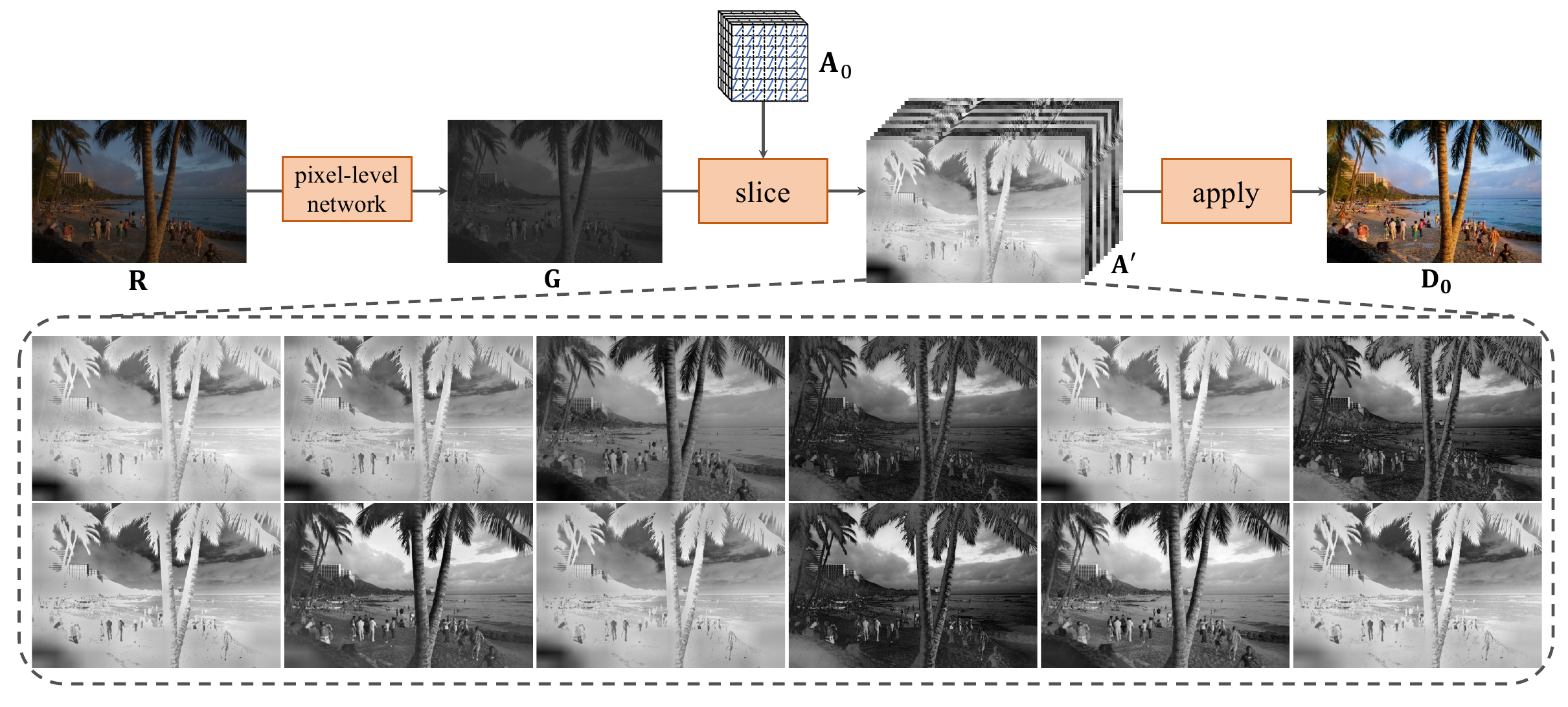}
    \vspace{-18pt}
    \caption{More details about the affine bilateral grid. By passing the input image $\mathimg{R}\in \mathbb{R}^{ H\times W\times 3}$ into a pixel-level network which is formulated as~\equref{eq:pixel}, the guidance map $\mathimg{G}\in \mathbb{R}^{ H\times W\times 1}$ can be obtained. Given the position and intensity indicated by $\mathimg{G}$. the operation of slicing performs a per-pixel lookup in $\mathimg{A}_0\in \mathbb{R}^{ H_{grid}\times W_{grid}\times D \times 12} $, and retrieves the sliced $\mathimg{A}'\in \mathbb{R}^{ H\times W\times 1 \times 12}  $. By applying the matrix multiply between the affine matrics and original color for each pixel in $\mathimg{R} $, the final result $\mathimg{D}_0 $ is obtained. The bottom two rows show the visual results of the sliced $\mathimg{A}'$. We can see that nearby pixels with similar intensities tend to retrieve similar affine matrics. These nearby pixels still maintain similar colors in the final output after the matrix multiplication, thus avoiding texture distortion.}
    \label{fig:sup_abg}
    \vspace{-6mm}
\end{figure*}

\section{Details about Affine Bilateral Grid}
\label{sec:abg}
In the main paper, 
we introduce the affine bilateral grid adopted in our~\ourmethod.
Here, we provide more details and visual results about it in~\figref{fig:sup_abg}.
The pixel-level network is similar to HDRNet~\cite{gharbi2017deep}.
It performs a pointwise nonlinear transformation to each pixel in the low-quality input $\mathimg{R}\in \mathbb{R}^{ H \times W \times 3}$ to obtain the guidance map $\mathimg{G}\in \mathbb{R}^{ H \times W \times 1}$,
which can be defined as
\begin{equation}
\label{eq:pixel}
    \mathimg{G}_\mathvec{p} = \mathvec{b} + \sum \rho (\mathmat{W} \mathimg{R}_\mathvec{p} + \mathvec{b'}),
\end{equation}
where $\mathvec{p}$ denotes the pixel position, and $\mathimg{G}_\mathvec{p}\in \mathbb{R}^{ 1 \times 1}$ and $\mathimg{R}_\mathvec{p}\in \mathbb{R}^{ 3 \times 1}$ denote the pixel value of $\mathimg{G}$ and $\mathimg{R}$ at position $\mathvec{p}$.
$W$ is a 3 × 3 color transformation matrix, which is initialized as an identical transformation.
$\sum$ is the summation operation along the channel dimension.
$\mathvec{b}\in \mathbb{R}^{ 1 \times 1}$ and $\mathvec{b}'\in \mathbb{R}^{ 3 \times 1}$ are the biases.
$\rho$ represents the nonlinear function including a sum of 16 scaled ReLU functions,
which can be written as
\begin{equation}
    \rho(\mathvec{x}) = \sum_{i=0}^{15}\mathvec{t}_i \text{max}(\mathvec{x} - \mathvec{n}_i,0),
\end{equation}
where $\mathvec{x}$ is the input vector.
$\mathvec{t}$ and $\mathvec{n}$ are the scalars and thresholds, which share the same shape with $\mathvec{x}$.
$\mathmat{W} $, $\mathvec{b} $, $\mathvec{b'} $, $\mathvec{t} $ and $\mathvec{n} $ are all learnable parameters and learned jointly with the other network parameters.

After obtaining the affine bilateral grid $\mathimg{A}_0 $, which is output by the underlying U-Net and restores a $3\times 4$ affine matrix in each grid,
we utilize the position and the intensity guided by $\mathimg{G}$ to perform a per-pixel lookup in $\mathimg{A}_0$.
Then the sliced $\mathimg{A}' $,
which contains the $3\times 4$ affine matrix for each pixel,
can be retrieved through trilinear interpolation.
We also show the visual results of the guidance map $\mathimg{G}$ and the sliced $\mathimg{A}'$ in~\figref{fig:sup_abg}.
We can see that the pixel-level network learns a gray-scale guidance map indicating the intensity for each pixel.
%
%
As shown in~\figref{fig:sup_abg},
due to the nature of the bilateral grid,
nearby pixels with similar intensities tend to retrieve similar affine matrics.
Thus after the matrix multiplication, 
these nearby pixels still maintain similar colors in the final output.
This is why the affine bilateral grid can overcome the texture distortion.

\begin{figure*}
    \centering
    \includegraphics[width=\linewidth]{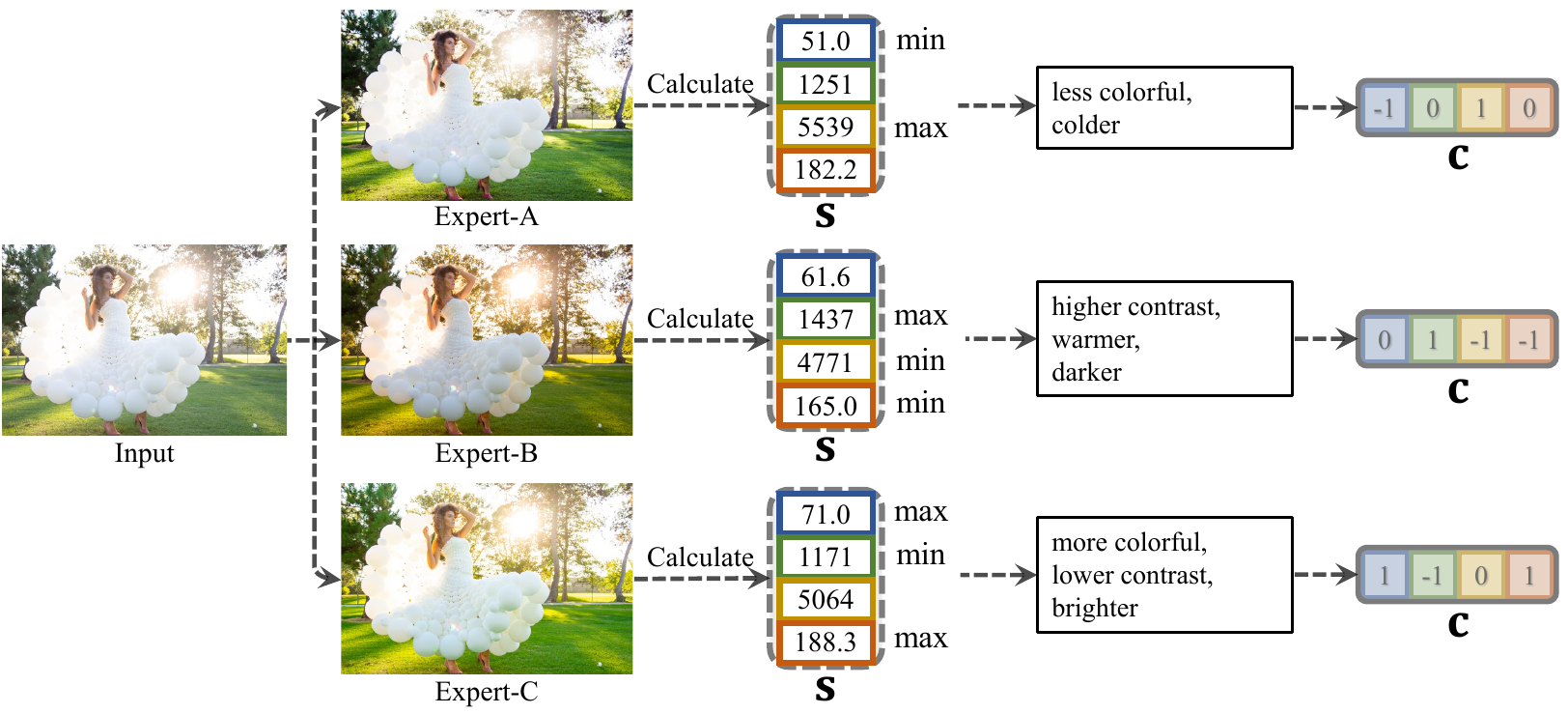}
    \caption{Example of constructing image-condition pairs. Given the ground truth (GTs) for the same low-quality image, we first calculate their scores $\mathvec{s}$. By comparing $\mathvec{s}_i$ for each contribute ($i$), corresponding labels are given to the GT with the highest or lowest $\mathvec{s}_i$. The condition $\mathvec{c}$ is obtained according to the labels for each GT. These image-condition pairs are used during both training and evaluation. During training, these conditions are fed into the intermediate layers of the backbone U-Net via a cross-attention layer, which is illustrated in~\secref{sec:net}. During evaluation, we provide the model with the pre-calculated condition of GT styles, thus making it produce corresponding output for evaluation.}
    \label{fig:sup_cons}
    \vspace{-5mm}
\end{figure*}

\section{Details about Constructing Image-condition Pairs}
\label{sec:cl}
Current datasets for retouching consist of results retouched by different experts.
However, 
they lack a description of the retouching style for each result.
In order to provide the users with an understandable editing mechanism,
we describe the retouching style from the aspects of four image attributes (colorfulness, contrast, color temperature, and brightness).
Therefore, 
we utilize the commonly used measure tools to evaluate each retouching result from four attributes (colorfulness, contrast, temperature, brightness),
and meanwhile construct the image-condition pairs.
These image-condition pairs are used during both training and evaluation.
During training, these conditions are fed into the intermediate layers of the backbone U-Net via a cross-attention layer, which is illustrated in~\secref{sec:net}.
Training with such image-condition pairs under the reconstruction supervision enables the model to sample various styles according to $\mathvec{c}$.
During evaluation, we provide the model with the pre-calculated condition of GT styles,
thus making it produce corresponding output for evaluation.

For a better explanation of the following parts, we define some
symbols here.
Let $\mathimg{I}$ represent the image to be scored, which is coded in the RGB colour space.
The three channels of $\mathimg{I}\in \mathbb{R}^{ H \times W \times 3}$ can be denoted as $\mathimg{R}\in \mathbb{R}^{ H \times W \times 1}$, $\mathimg{G}\in \mathbb{R}^{ H \times W \times 1}$, and $\mathimg{B}\in \mathbb{R}^{ H \times W \times 1}$.
%
During the calculation, we may frequently utilize the operation of calculating the mean value and the standard deviation of some one-channel images on the spatial level.
Thus we also define the two functions here.
Let $\mathimg{M}$ represent arbitrary one-channel image,
and the function $\mu$ is defined to calculate the mean value of $\mathimg{M}$ as follows
\begin{equation}
    \mu(\mathimg{M}) = \frac{\sum_{\mathvec{p}\in \mathimg{M} }\mathimg{M}_\mathvec{p}}{N},
\end{equation}
where $N$ is the number of the pixels in $\mathimg{M}$ and $\mathvec{p}$ denotes the pixel position in $\mathimg{M}$.
Then the function $\sigma$ is defined to calculate the standard deviation as
\begin{equation}
    \sigma(\mathimg{M}) = \sqrt{\frac{1}{N} \sum_{\mathvec{p}\in\mathimg{M} } (\mathimg{M}_\mathvec{p} - \mu(\mathimg{M}))^2  }.
\end{equation}

The calculated score vector $\mathvec{s}$ is denoted as $\mathvec{s}=[\mathvec{s}_1,\mathvec{s}_2,\mathvec{s}_3,\mathvec{s}_4]$,
which corresponds to the score of colorfulness, contrast, color temperature, and brightness respectively.
Specifically,
for measuring the overall colorfulness of the image,
we adopt the colorful metric proposed in~\cite{colorfulness},
which can be formulated as :
\begin{equation}
    \mathvec{s}_{1} = \sqrt{[\sigma(\mathimg{H})]^2 + [\sigma(\mathimg{U})]^2}+ 0.3\sqrt{[\mu(\mathimg{H})]^2 + [\mu(\mathimg{U})]^2},
\end{equation}
where $\mathimg{H} = \mathimg{R}- \mathimg{G}$, $\mathimg{U} = \frac{1}{2}(\mathimg{R}+\mathimg{G})-\mathimg{B}$.
%

Since contrast represents the difference in color that makes images more clear,
we utilize the sum of squared differences as the measurement of the image contrast:
\begin{equation}
    \mathvec{s}_{2} = \sum_{\mathvec{j}\in \mathimg{I}}\frac{\sum_{\mathvec{i} \in \omega}(\mathimg{I}_\mathvec{i} - \mathimg{I}_\mathvec{j})^2}{N_\omega},
\end{equation}
where $\mathvec{i}$ and $\mathvec{j}$ denote the pixel position,
$\omega$ represents the neighbors to every pixel, and $N_\omega$ is the number of the neighbor pixels.
In our setting, we adopt 4-neighbours of each pixel for calculation.

As for the color temperature,
we adopt Correlated Color Temperature (CCT)~\cite{hernandez1999calculating}, which is measured in degrees Kelvin (K).
It firstly involves transforming the pixel value from sRGB colorspace to XYZ tristimulus value defined by CIE 1931.
Thus the three channels are transformed as $\mathimg{X}$, $\mathimg{Y}$, and $\mathimg{Z}$.
Then we can obtain the value of $x$ and $y$ by
\begin{equation}
    x = \frac{\mu(\mathimg{X})}{\mu(\mathimg{X}) + \mu(\mathimg{Y}) + \mu(\mathimg{Z})}, y = \frac{\mu(\mathimg{Y})}{\mu(\mathimg{X}) + \mu(\mathimg{Y}) + \mu(\mathimg{Z})}.
\end{equation}
The formula for calculating CCT can be written as
\begin{equation}
    \begin{split}
        \mathvec{s}_{3} = 62453.8 e ^ {-n / 0.9} + 28.7 e ^ {-n / 0.2}+ 0.00004 e ^ {-n / 0.1}  - 949.9,  
    \end{split}
\end{equation}
where $n = \frac{x-0.3366}{y - 0.1735}$.

For brightness,
we use the perceived brightness,
which can be calculated by:
\begin{equation}
    \mathvec{s}_{4} = \sqrt{0.241[\mu(\mathimg{R})]^2 + 0.691[\mu(\mathimg{G})]^2 + 0.068[\mu(\mathimg{B})]^2}.
\end{equation}

With these measurements, we can construct the image-condition pairs, where an example of this process is shown in~\figref{fig:sup_cons}.
For each low-quality image,
the dataset provides several ground truth (GTs) retouched by different experts.
Among these GTs for the same input,
we calculate $\mathvec{s}$ for each one,
which is denoted as $\mathvec{s}=[\mathvec{s}_1,\mathvec{s}_2,\mathvec{s}_3,\mathvec{s}_4]$, and each $\mathvec{s}_i$ correspond to one attribute.
For each attribute ($i$),
we compare $\mathvec{s}_i$ of these GTs.
We set the corresponding condition $\mathvec{c}_i$ of the GT with the highest(lowest) $\mathvec{s}_i$ as 1(-1), otherwise $\mathvec{c}_i$ is set to 0.


\section{Evaluation Details}
\label{sec:test}
As mentioned in the main paper,
we compare our \ourmethod~with other methods proposed for retouching on MIT-Adobe FiveK and PPR10K datasets.
For MIT-Adobe FiveK, the deterministic methods which can only produce a single retouching style for each input are trained and evaluated on the Expert-C subset.
These methods include HDRNet~\cite{gharbi2017deep},
DeepUPE~\cite{wang2019underexposed},
CURL~\cite{moran2021curl},
DeepLPF~\cite{moran2020deeplpf},
3DLUT~\cite{zeng2020learning},
CSRNet~\cite{he2020conditional},
and RSFNet~\cite{ouyang2023rsfnet}.
The other type of methods supports multiple retouching styles,
which includes PIENet~\cite{kim2020pienet}, StarEnhancer~\cite{song2021starenhancer}, TSFlow~\cite{wang2022learning}, and our~\ourmethod.
%
PIENet~\cite{kim2020pienet}, TSFlow~\cite{wang2022learning}, StarEnhancer~\cite{song2021starenhancer}, and our \ourmethod~are trained with the mixture of five subsets (Expert-A/B/C/D/E).
During inference,
for PIENet~\cite{kim2020pienet},
we directly evaluate the corresponding results for five subsets provided by the author.
For StarEnhancer~\cite{song2021starenhancer} and TSFlow~\cite{wang2022learning},
to generate the corresponding prediction for each subset,
the style embedding, which represents the overall retouching style of the expert, is extracted following the practice in their original paper and sent into their model.
Our \ourmethod~produces the corresponding prediction for each subset with the constructed condition, which is introduced in~\figref{fig:sup_cons}.

For PPR10K dataset,
four deterministic methods including HDRNet~\cite{gharbi2017deep}, CSRNet~\cite{he2020conditional}, 3DLUT~\cite{zeng2020learning}, and RSFNet~\cite{ouyang2023rsfnet} are compared.
%
These methods train three models on three subsets respectively,
and evaluate the three models on the corresponding subset.
Our \ourmethod~is trained with the mixture of the three subsets (Expert-A/B/C) simultaneously and produces the corresponding prediction for each subset with the constructed condition, which is introduced in~\figref{fig:sup_cons}.
%

%
For the evaluation with FID~\cite{heusel2017gans},
the reference images are set as a mixture of subsets to evaluate the similarity between the distribution of the predictions and that of expert-retouched GTs.
For MIT-Adobe FiveK dataset,
we only calculate FID for methods supporting additional styles including PIENet~\cite{kim2020pienet}, StarEnhancer~\cite{song2021starenhancer}, and TSFlow~\cite{wang2022learning},
where the predictions for five subsets are combined to calculate the FID value.
For HDRNet~\cite{gharbi2017deep}, CSRNet~\cite{he2020conditional}, 3DLUT~\cite{zeng2020learning} and RSFNet~\cite{ouyang2023rsfnet} on PPR10K dataset,
we put the results produced by three models together to calculate FID.

\vspace{-3mm}
\begin{figure}[h]
  \centering
  \begin{minipage}{0.6\linewidth}
    \includegraphics[width=\linewidth]{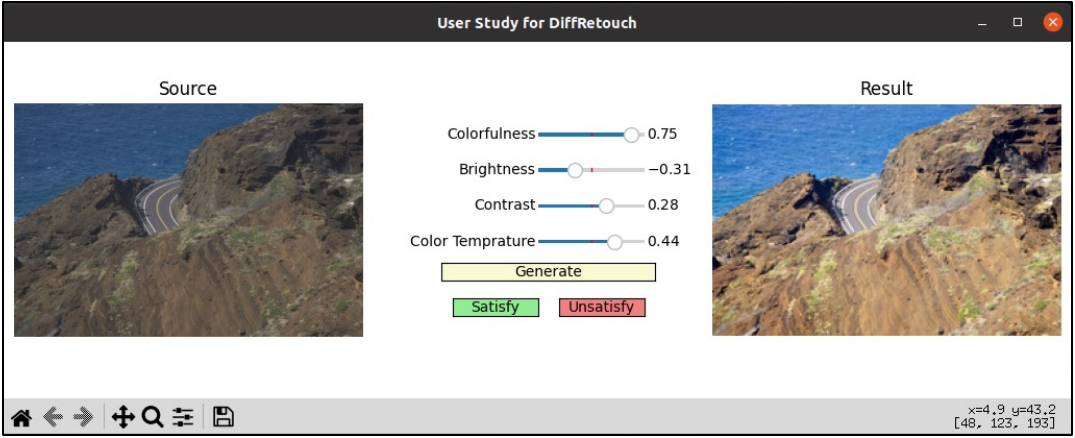}
    \vspace{-6mm}
    \caption{UI of user study.}
    \label{fig:us_ui}
  \end{minipage}
  \hfill
  \begin{minipage}{0.36\linewidth}
    \begin{tabular}{c|c}
    \toprule
    Samples         & 600   \\
    \midrule
    Avg. Operations & 2.67  \\
    \midrule
    Avg. Time       & 13.70s  \\
    \midrule
    Failure         & 4       \\
    \bottomrule
    \end{tabular}
    \vspace{-3mm}
    \tabcaption{Results of user study.}
    \label{tab:us_result}
  \end{minipage}
  \vspace{-9mm}
\end{figure}
\section{User Study}
\label{sec:usa}
To further show the advantages of our method,
we conduct another user study.
We invited 30 volunteers, and randomly chose 20 images.
UI is shown in \figref{fig:us_ui}, where the input is shown on the left.
Users can adjust the four attributes,
and the output will appear on the right after clicking `Generate'.
The user can choose `Satisfy' or `Unsatisfy' based on their preferences.
If click `Unsatisfy', users will adjust the attributes and generate the results again.
We recorded the number of times needed to click `Generate' until satisfied as the number of operations.
If it exceeds 15 times, it is considered a failure and recorded as 15.
As shown in \tabref{tab:us_result},
%
despite different preferences, 
our method enables almost all users to achieve `Satisfy' in 3 operations,
showing its ability to meet various personalization needs.

\section{Analysis of the Attributes}
\label{sec:att}
Four adjustable attributes are provided to the users to control the output within the fine-retouched distribution.
For example, 
when $\mathvec{c}_i$ is set to 1,
this will make the output more colorful, higher contrast, cold temperature, or brighter one among the fine-retouched results,
while the output will be less colorful, lower contrast, warm temperature, or darker one when $\mathvec{c}_i$ equals -1.

\begin{figure}[t]
    \centering
    \includegraphics[width=0.8\linewidth]{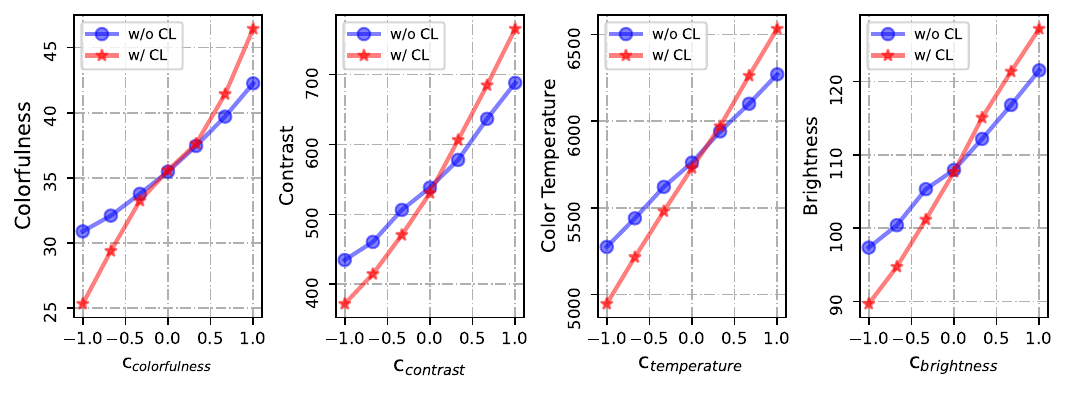}
    \vspace{-3mm}
    \caption{Evaluation of contrastive learning.}
    \label{fig:cl_step}
    \vspace{-3mm}
\end{figure}

In Table 5 of the main paper,
we compare the adjustable range w/ and w/o contrastive learning.
Here, we make it clearer by plotting the graphs of four-attribute scores vs. control-parameter values in \figref{fig:cl_step}.
We can see that with contrastive learning, 
our method has a larger adjustable range for each attribute, suggesting the effectiveness of contrastive learning.
On the other hand,
the adjustment process is almost linear, 
which demonstrates the control parameters are mapped properly, 
and showcases the effect of the attributes learning.

We also conduct experiments to evaluate the independence of these attribute controls.
We change one single attribute $\mathvec{c}_i$ from maximum (1) to minimum (-1), while others keep static to calculate each attribute score $\mathvec{s}_i$ of the output.
The adjustable range for each attribute is determined by the changes in the corresponding scores ($\mathvec{s}_i$) of outputs,
and a larger adjustable range means a more significant influence brought
by adjusting the coefficient.
The results are shown in~\tabref{tab:score}.
We can see that the adjustable range for each attribute ($i$) is significantly larger when changing the corresponding attribute ($\mathvec{c}_i$) than when changing other attributes.
This proves that our method has almost decoupled the control of different attributes.

\setlength{\tabcolsep}{1.6pt}
\begin{table}[t]
\centering
\newcommand{\kcl}{$^{\dag}$~}
\newcommand{\kcg}{$^{\S}$~}
\begin{tabular}{c|c|c|c|c}
    \toprule
    \diagbox{$\mathvec{c}_i$}{$\mathvec{s}_i$}    & Colorfulness$\uparrow$          & Contrast$\uparrow$      &  Color  Temp.$\uparrow$     & Brightness$\uparrow$  \\
    \midrule
    Colorfulness  &   \best{20.60}            &   102.9              &   837.4              &  5.09      \\
    Contrast    &   6.16           &    \best{297.3}              &   324.7              &  7.85       \\      
    Color Temp.    &   3.78           &    41.9              &   \best{1572.9}              &  5.43       \\    
    Brightness    &   3.91           &    76.2              &   614.4              &  \best{35.22}       \\    
    \bottomrule
\end{tabular}
\caption{Adjustable range for each attribute. The adjustable range is determined by the changes in the scores ($\mathvec{s}_i$) when changing one single attribute $\mathvec{c}_i$ from maximum (1) to minimum (-1).}
\label{tab:score}
\vspace{-6mm}
\end{table}

\section{More Visual Results for Total Number of Sampling Steps}
\label{sec:total}
\begin{figure}[h]
    \centering
    \includegraphics[width=0.96\linewidth]{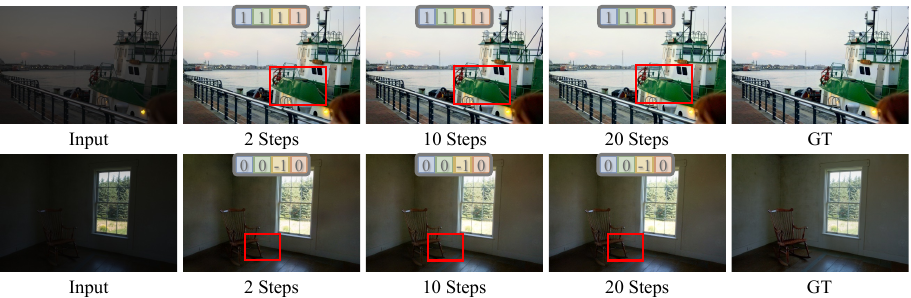}
    \caption{Visual results when adopting different total numbers of sampling steps. `$n$ Steps' represents that the total number of sampling steps is set as $n$. The input condition $\mathvec{c}$ is shown at the top of each result generated by DiffRetouch. Focusing on the content within the red box can see the difference in brightness and color.}
    \label{fig:sup_step1}
    \vspace{-6mm}
\end{figure}

To investigate the effect of the sampling steps,
we carry out two experiments on our adopted improved DDPM~\cite{chung2022improving}.
In the first experiments, 
we set different total numbers of sampling steps like 2, 10, and 20,
and visualize these results,
which is shown in~\figref{fig:sup_step1}.
We can see that,
when the total number of sampling steps is small like 2,
the output may encounter the problems of underexposure and color shift.
With the total number of sampling steps increasing,
these problems are improved.
When the number is set as 20, which is the default number in our main paper,
we can obtain satisfactory results with brighter conditions and less color shift.

\begin{figure}[ht]
    \centering
    \includegraphics[width=0.96\linewidth]{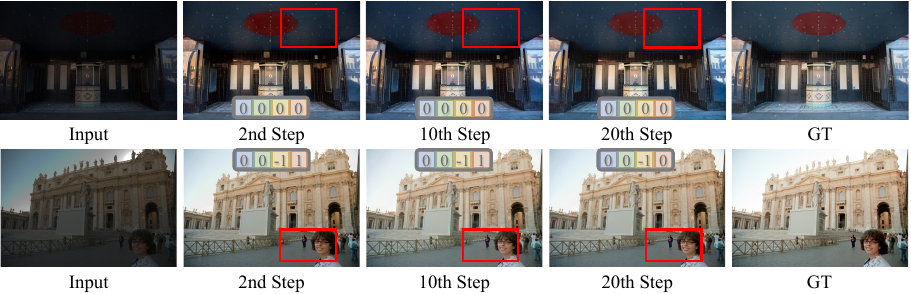}
    \caption{Visual results of the intermediate results during the sampling process. `$n$-th Step' represents that the results are obtained when the sampling process reaches the $n$-th step. The total number of the sampling steps is set as 20. The input condition $\mathvec{c}$ is shown at the top of each result generated by DiffRetouch. Focusing on the content within the red box can see the difference in brightness and color.}
    \label{fig:sup_step2}
    \vspace{-6mm}
\end{figure}

\section{More Visual Results for Intermediate Results during Sampling Process}
\label{sec:inter}
The second experiment is carried out when the total number of sampling steps is set as 20,
and we visualize the intermediate results during the sampling process.
For example,
when it comes to the 2nd sampling step,
we can obtain the intermediate bilateral grid $\mathimg{A}_2$.
With the operation of slicing and applying,
we can obtain the intermediate result $\mathimg{D}_2$.
These intermediate results of the 2nd, 10th, and 20th steps are shown in~\figref{fig:sup_step2}.
We can see that at the early stage of sampling,
our \ourmethod~pays more attention to the overall characteristic of the image like brightness.
In the later stage,
our \ourmethod~plays the role of refinement like color correction.

\section{Limitations}
\label{sec:limit}
\begin{figure}[t]
    \centering
    \includegraphics[width=0.8\linewidth]{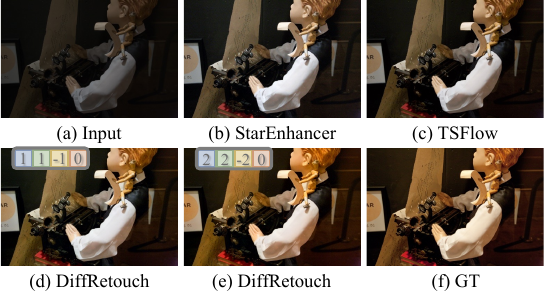}
    \caption{Limitations of~\ourmethod. (a) is the input image, and (b) and (c) are the results of StarEnhancer~\cite{song2021starenhancer} and TSFlow~\cite{wang2022learning}. Limited by their network architecture, they tend to learn the average style of the experts, and are hard to achieve the style of one specific expert. (d) and (e) are the results of our~\ourmethod, which differ in the input condition $\mathvec{c}$. (f) is the Ground Truth (GT). The result of (d) can hardly realize the style of (f). After adjusting the condition $\mathvec{c}$, the result of (e) can be closer to GT.}
    \label{fig:sup_limit}
    \vspace{-5mm}
\end{figure}

Although our \ourmethod~can satisfy various aesthetic preferences in most cases,
there are still some failure cases we encounter during evaluation.
\figref{fig:sup_limit} shows such a failure case.
Limited by their network architecture,
TSFlow~\cite{wang2022learning} and StarEnhancer~\cite{song2021starenhancer} tend to learn the average style of the experts and are hard to achieve the style of one specific expert.
As shown in~\figref{fig:sup_limit} (d),
our \ourmethod~performs better than them, but still fails in this extreme case for $\mathvec{c}_i$ within $[-1,1]$.
Note that the conditions of~\figref{fig:sup_limit} (d) are pre-calculated by the rule introduced in~\secref{sec:cl} to make our model produce corresponding output for evaluation, 
which is constrained within $[-1,1]$.
However, 
when used in practice, 
these conditions $\mathvec{c}_i$ are provided by users,
and users are allowed to adjust $\mathvec{c}_i$ freely to suit their preferences.
As shown in~\figref{fig:sup_limit} (e),
when the provided conditions are twice those of~\figref{fig:sup_limit} (d),
the output can be closer to the desired result.

\begin{figure}[h]
    \centering
    \includegraphics[width=0.75\linewidth]{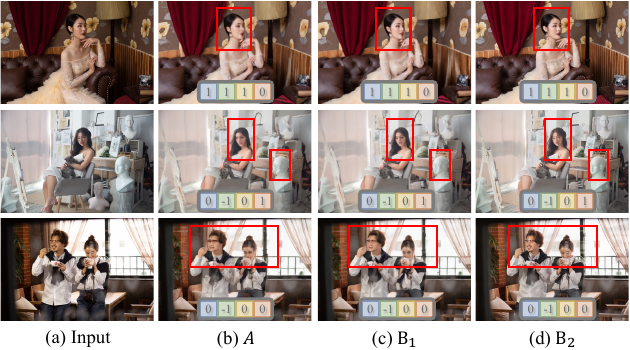}
    \caption{More Visual Results for Affine Bilateral Grid. `$\rm A$' represents the baseline model which suffers from severe texture distortion. `$\rm B_1$' and `$\rm B_2$' represent the baseline model equipped with CFW~\cite{wang2023exploiting} and affine bilateral grid. CFW can somewhat alleviates the severe texture distortion, while the affine bilateral grid well solves this problem.}
    \label{fig:sup_ablation}
\end{figure}

\begin{figure}[t]
    \centering
    \includegraphics[width=0.75\linewidth]{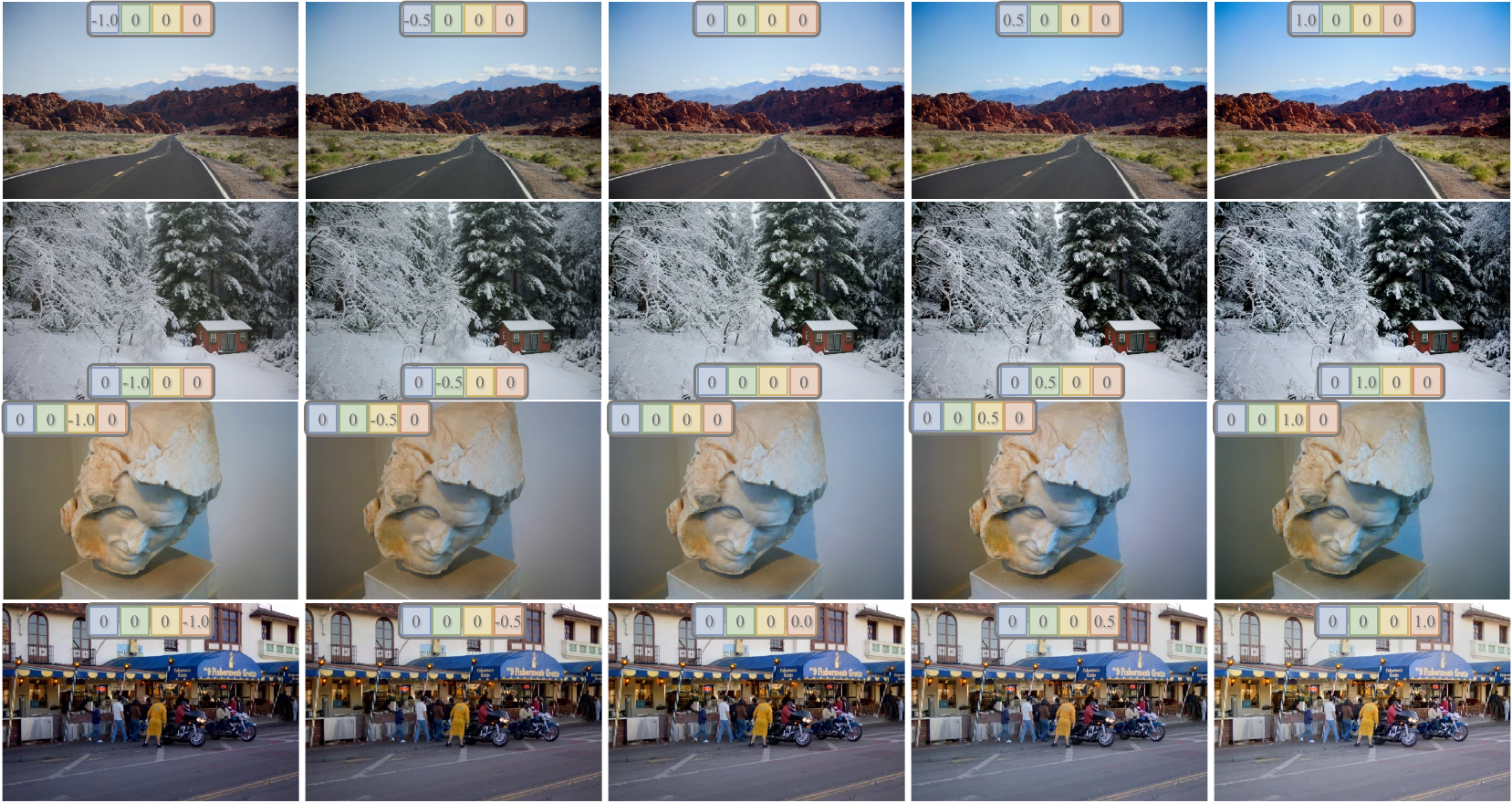}
    \captionof{figure}{Results with a continuum of variations (colorfulness, contrast, color temperature, and brightness) from -1 to 1 with a step of 0.5.}
    \label{fig:continous}
    \vspace{-4mm}
\end{figure}

\section{More Visual Results for Affine Bilateral Grid}
\label{sec:ablation}
In \figref{fig:sup_ablation},
we show more visual results to demonstrate the effectiveness of the affine bilateral grid.
Our intention of introducing the affine bilateral grid is to solve the texture distortion existing in our baseline model, which is shown in \figref{fig:sup_ablation} (b).
Motivated by StableSR~\cite{wang2023exploiting},
the controllable feature wrapping (CFW) module is adopted,
which extracts the intermediate features from the encoder to modulate the decoder features.
As shown in \figref{fig:sup_ablation} (c),
the severe texture distortion can be somewhat alleviated.
However, it cannot handle complex structures like human faces.
With the introduction of the affine bilateral grid,
this problem can be well solved and original details are reserved,
which is shown in \figref{fig:sup_ablation} (d).

\section{More Visual Results for Contrative Learning and Adjustable Range}
\label{sec:sup_cl}
\begin{figure}[ht]
    \centering
    \includegraphics[width=0.96\linewidth]{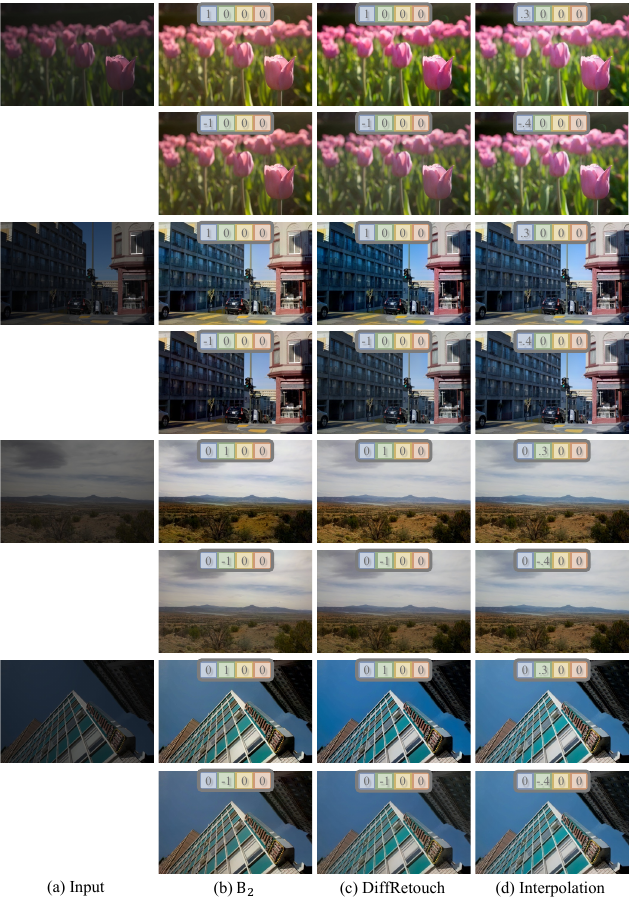}
    \caption{More Visual Results for Contrative Learning and Adjustable Range. $\rm B_2$ represents the baseline model equipped with the affine bilateral grid w/o contrastive learning scheme. (c)-(d) show the retouching styles generated by our \ourmethod~with uniformly varying conditions and only one attribute is changed at one time. The changed attributes are colorfulness and contrast in order, and each occupies four rows. The adjustable ranges are enlarged for these two attributes and our \ourmethod~can generate more intermediate styles between these extreme cases.}
    \label{fig:sup_cl1}
    \vspace{-5mm}
\end{figure}

\begin{figure}[ht]
    \centering
    \includegraphics[width=0.96\linewidth]{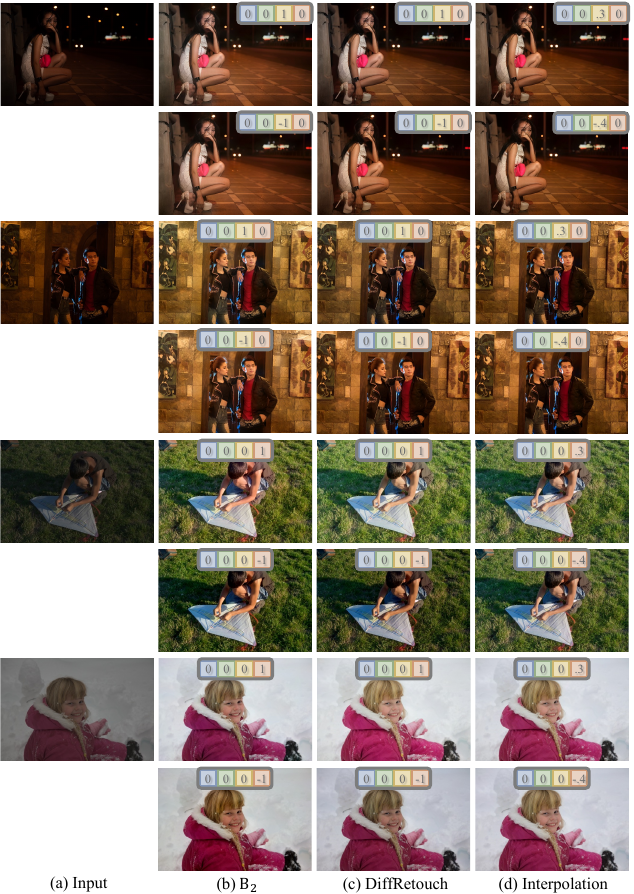}
    \caption{More Visual Results for Contrative Learning and Adjustable Range. $\rm B_2$ represents the baseline model equipped with the affine bilateral grid w/o contrastive learning scheme. (c)-(d) show the retouching styles generated by our \ourmethod~with uniformly varying conditions and only one attribute is changed at one time. The changed attributes are color temperature and brightness in order, and each occupies four rows. The adjustable ranges are enlarged for these two attributes and our \ourmethod~can generate more intermediate styles between these extreme cases.}
    \label{fig:sup_cl2}
    \vspace{-5mm}
\end{figure}

We provide more examples in \figref{fig:continous} with a step of 0.5.
We can see that as the coefficients increase, the images change smoothly and in the correct direction.
In \figref{fig:sup_cl1}~and \figref{fig:sup_cl2},
we show more visual results to demonstrate the effectiveness of the contrastive learning scheme.
First, for each attribute ($i$),
we set the corresponding condition $\mathvec{c}_i$ to 1 and -1, and the results generated by $\rm B_2$ (the baseline model equipped with the affine bilateral grid but w/o contrastive learning scheme) and the full model (\ourmethod) are shown in \figref{fig:sup_cl1}~and \figref{fig:sup_cl2} (c) and (d).
Each attribute occupies four rows of images.
We can see that with the implementation of the contrastive learning scheme,
when the conditions of extreme values are given, the results show more prominent feature \wrt corresponding attributes.
For example,
in the first four rows where the changed attribute is colorfulness,
the images in (c) are more/less colorful than in (b) with the same condition input.
The situation is similar for other attributes.
This also means that the adjustable ranges are enlarged for these four attributes and our \ourmethod~can generate more intermediate styles between these extreme cases.
Then \figref{fig:sup_cl1}~and \figref{fig:sup_cl2}~(c)-(d) show the retouching styles generated by our \ourmethod~with uniformly varying conditions.
We can see that the style changes continuously and in the correct direction for the corresponding attribute.
With the larger adjustable range and friendly editing mechanism,
users are allowed to customize their preferred style by adjusting the coefficients \wrt four attributes.

\section{More Visual Results for comparison with other SOTA methods}
\label{sec:sup_visual}
\begin{figure*}
    \centering
    \includegraphics[width=\linewidth]{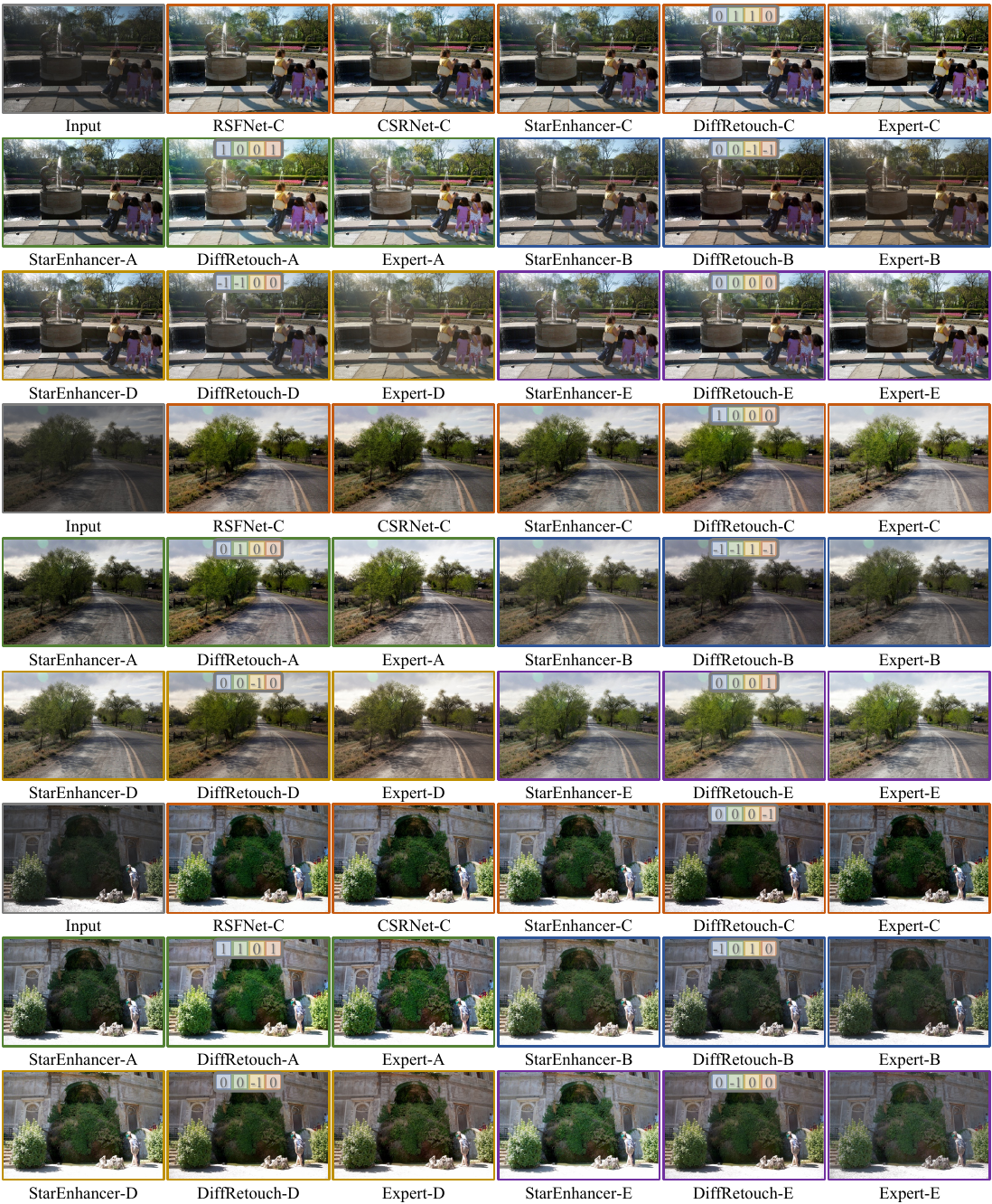}
    \caption{More qualitative comparison on MIT-Adobe FiveK with subsets retouched by five experts (A/B/C/D/E). The input condition $\mathvec{c}$ is shown at the top of each result generated by DiffRetouch.}
    \label{fig:sup_adobe1}
    \vspace{-5mm}
\end{figure*}

\begin{figure*}
    \centering
    \includegraphics[width=\linewidth]{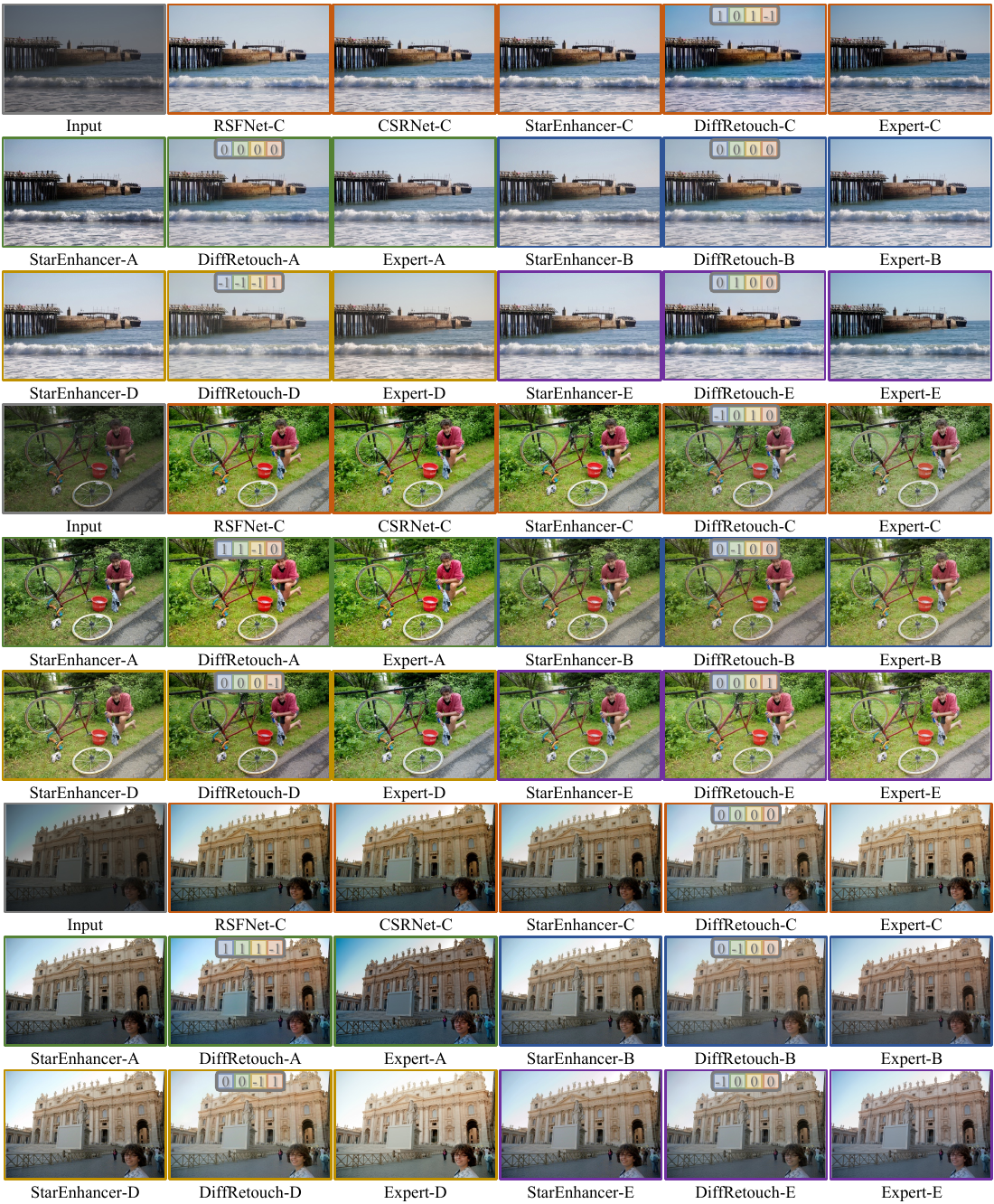}
    \caption{More qualitative comparison on MIT-Adobe FiveK with subsets retouched by five experts (A/B/C/D/E). The input condition $\mathvec{c}$ is shown at the top of each result generated by DiffRetouch.}
    \label{fig:sup_adobe2}
    \vspace{-5mm}
\end{figure*}

\begin{figure*}
    \centering
    \includegraphics[width=\linewidth]{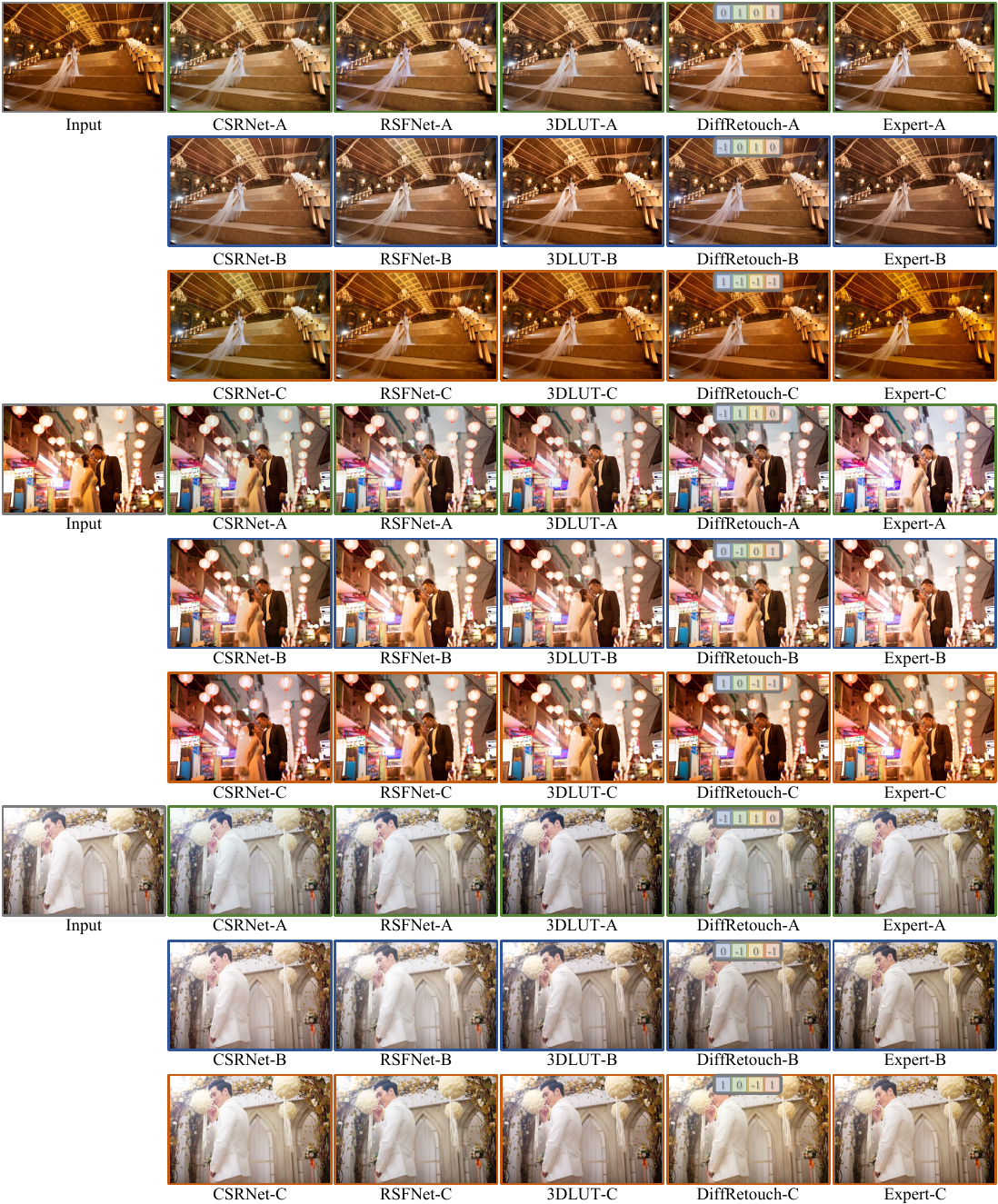}
    \caption{More visual results on PPR10K dataset with subsets retouched by three experts (A/B/C). The input condition $\mathvec{c}$ is shown at the top of each result generated by DiffRetouch.}
    \label{fig:sup_ppr1}
    \vspace{-5mm}
\end{figure*}

\begin{figure*}
    \centering
    \includegraphics[width=\linewidth]{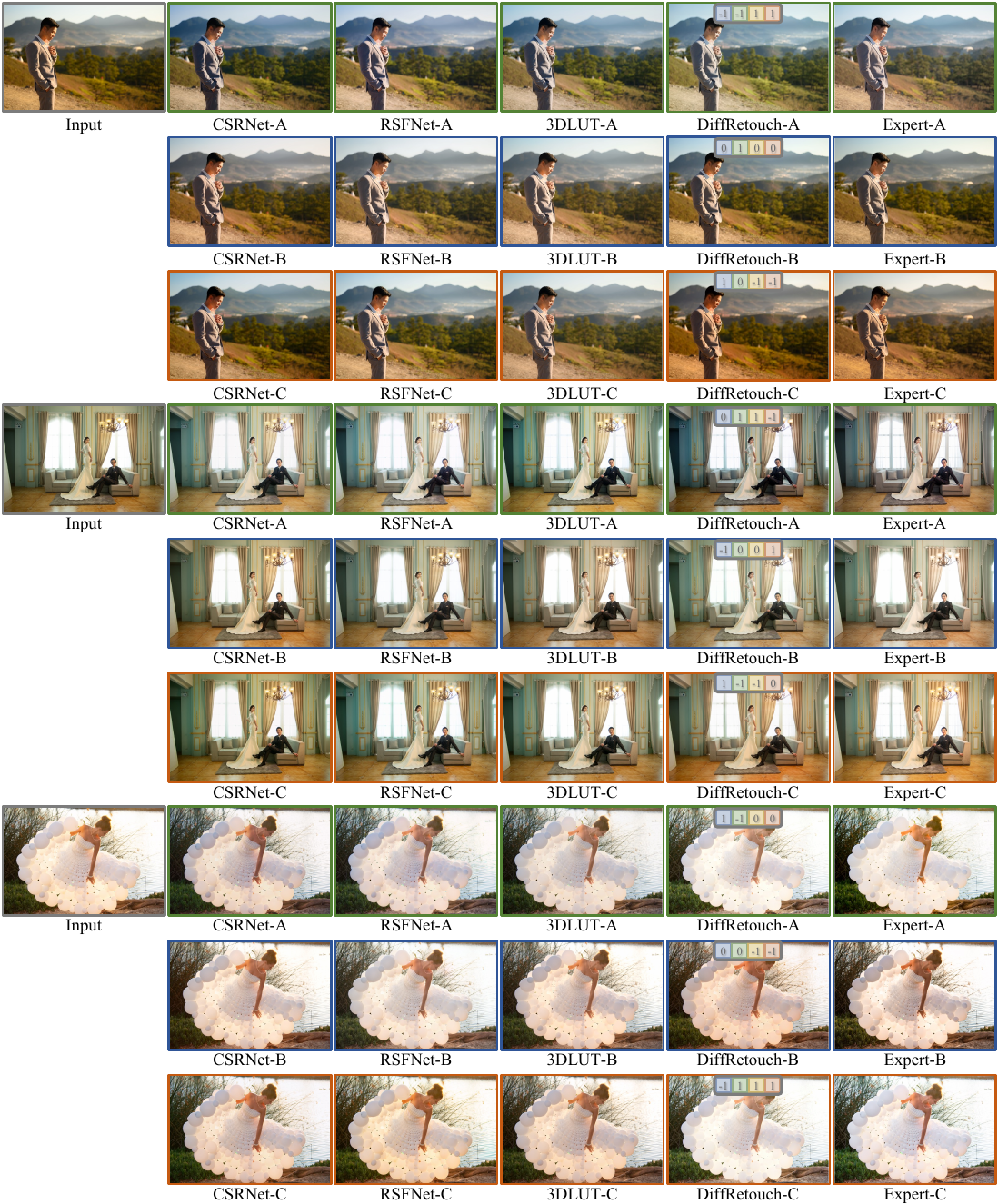}
    \caption{More visual results on PPR10K dataset with subsets retouched by three experts (A/B/C). The input condition $\mathvec{c}$ is shown at the top of each result generated by DiffRetouch.}
    \label{fig:sup_ppr2}
    \vspace{-5mm}
\end{figure*}

\figref{fig:sup_adobe1} and \figref{fig:sup_adobe2} show more qualitative comparisons against the state-of-the-art methods on MIT-Adobe FiveK dataset with subsets retouched by five experts (A/B/C/D/E).
The results generated by our \ourmethod~are more similar to the corresponding expert-retouched results.
\figref{fig:sup_ppr1} and \figref{fig:sup_ppr2} shows more visual results on PPR10K dataset with subsets retouched by three experts (A/B/C). 
By adjusting the attributes, our \ourmethod~can easily generate the results sharing similar styles with the expert-retouched results.

%
%

\clearpage
\end{document}